\documentclass[a4paper,fleqn]{cas-sc}

\usepackage[authoryear,longnamesfirst]{natbib}

\usepackage{amsmath,amssymb}
\usepackage{graphicx}
\usepackage{booktabs}
\usepackage{array}
\usepackage{tabularx}
\usepackage{makecell}
\usepackage{fvextra}

\usepackage{subcaption}

\begin{document}
\let\WriteBookmarks\relax
\def\floatpagepagefraction{1}
\def\textpagefraction{.001}

\shorttitle{A Multi-Agent LLM System for Automated Tuning}
\shortauthors{Moloshnikov et al.}
\title[mode = title]{A Multi-Agent LLM System for Automated Tuning of Specialized Neural Network Models}

\author[1]{Ivan Moloshnikov}[orcid=0000-0002-5000-7257]
\cormark[1]
\ead{ivan-rus@yandex.ru}
\credit{Conceptualization, Methodology, Software, Investigation,
Data curation, Writing -- original draft}

\author[1]{Dmitry Kunitsyn}[orcid=0009-0002-9640-6787]
\ead{Kunitsyn_DE@nrcki.ru}
\credit{Methodology, Investigation, Software,
Writing -- original draft}

\author[1,2]{Alexander Sboev}[orcid=0000-0002-6921-4133]
\credit{Conceptualization, Methodology, Investigation,
Data curation, Supervision, Validation,
Writing -- original draft}
\cormark[2]
\ead{sag111@mail.ru}

\author[1]{Maksim Skorokhodov}[orcid=0000-0001-9399-9332]
\ead{Skorokhodov_MS@nrcki.ru}
\credit{Software, Validation, Investigation, Visualization,
Writing -- original draft}

\author[1,3]{Roman Rybka}[orcid=0000-0002-5595-6398]
\ead{Rybka_RB@nrcki.ru}
\credit{Supervision, Project administration}

\author[4]{Yuanchao Liu}[orcid=0009-0007-0430-7658]
\ead{ycliuharbin@163.com}
\credit{Formal analysis, Methodology}

\affiliation[1]{
  organization={National Research Centre ``Kurchatov Institute''},
  addressline={1 Akademika Kurchatova Square},
  city={Moscow},
  postcode={123182},
  state={Moscow},
  country={Russia}
}

\affiliation[2]{
  organization={National Research Nuclear University MEPhI},
  addressline={31 Kashirskoye Shosse},
  city={Moscow},
  postcode={115409},
  state={Moscow},
  country={Russia}
}

\affiliation[3]{
  organization={Moscow Institute of Physics and Technology},
  addressline={9 Institutskiy Lane},
  city={Dolgoprudny},
  postcode={141701},
  state={Moscow},
  country={Russia}
}

\affiliation[4]{
  organization={Harbin Institute of Technology},
  addressline={No. 92 West Dazhi Street, Nangang District},
  city={Harbin},
  postcode={150001},
  state={Heilongjiang},
  country={China}
}

\cortext[1]{Corresponding author}
\cortext[2]{Principal corresponding author}

\begin{abstract}
Many machine learning approaches are two-level: an expert selects a class of methods, within this class the choice of key components is critical and usually manual. This paper addresses the automatic tuning of these components and proposes a framework transferable across models of different nature. The multi-agent large language model (LLM) framework has three roles---analyst, coder, and recorder~---with an experiment log and a sandboxed execution environment. Unlike systems that optimize the entire ML script or search architectures in a discrete space, the framework operates over an open but semantically narrow space: the agent generates code only for the specified components, leaving the overall network structure unchanged. The approach is validated in two directions: tuning a classification spiking neural network with correlation learning, and solving high-order differential equations with Physics-Informed Neural Networks (PINN). Across five tasks---spiking classification on MNIST and Digits, and boundary-value problems for the generalized nonlinear Schr\"odinger equation (GNLSE) and Kuramoto--Sivashinsky (KS) equations---the agent outperforms the baselines: classification accuracy is 13 percentage points (MNIST) and 6 percentage points (Digits) higher; the error against the analytical single-soliton solution is reduced approximately 2491-fold; the invariant-preservation error drops approximately 325-fold in the selected GNLSE run with a Gaussian perturbation and by more than 100-fold on average over repeated initializations; and the mass drift drops 14-fold for the KS problem with a Gaussian perturbation. The framework thus tunes key components on tasks of different nature without changing its internal structure.  
\end{abstract}

\begin{highlights}
    \item A multi-agent large language model framework tunes functional blocks of specialized neural networks.
    \item The same framework transfers across five spiking classification and physics-informed neural network tasks.
    \item MNIST accuracy rises +13 pp; selected GNLSE errors drop up to 2500x and 325x.
    \item Detailed prompts substantially improve performance of less capable large language models.
\end{highlights}

\begin{keywords}
Large language models
\sep Multi-agent systems
\sep Automated machine learning
\sep Physics-informed neural networks
\sep Spiking neural networks
\end{keywords}

\maketitle

\section{Introduction}
\label{sec:intro}

Large language models (LLMs) have become the foundation of a new class of tools~---agent systems that can not only respond to queries but also plan actions, use tools, execute code, and accumulate experience from previous steps~\cite{xi2023rise,yao2023react}. 
Such agents are already used in programming, ML engineering, and individual scientific disciplines~\cite{jimenez2024swebench,chan2024mlebench,lu2024scientist,schmidgall2025agent}; however, their use as a meta-optimizer for specialized neural network models remains understudied. A detailed review of related work is given in Section~\ref{sec:related}.
 
The quality of specialized neural network models depends strongly on the task: the representation of inputs and outputs, the method for decoding internal activity, the balance of loss-function terms, and other components must be selected anew for each problem, and no universal selection rule exists. In practice, this selection is performed manually, which makes the process labor-intensive and poorly reproducible. At the same time, models of different types can be decomposed into similar modules: input encoding and preprocessing, a trainable feature transformation, a decoding or output transformation, and a loss function. The tuning task is therefore reduced not to redesigning the entire network for every new problem but to modifying its individual modules, which enables a single tuning procedure applicable to models of different types.
 
Studies in which LLM agents are used as a general tool for automatic tuning of individual functional blocks of a model with a fixed architecture and training procedure are almost absent from the open literature (see Section~\ref{sec:related}). Existing systems focus either on programming tasks and the general ML cycle or on constructing a network from a textual task description. In contrast, the proposed framework does not build a network but modifies its predefined blocks using a single agent interaction pattern that does not depend on the particular application task.
 
The main contributions are as follows.
\begin{enumerate}
    \item We propose a multi-agent LLM framework for the automatic tuning of individual functional blocks of a neural network model (Section~\ref{sec:agent}).
    \item We show that the same framework transfers across five tasks with different types of information representation without changes to its internal structure (Section~\ref{sec:results}, Table~\ref{tab:summary}).
    \item We show that a detailed prompt improves the performance of less capable LLMs, bringing it closer to that of more advanced models (Section~\ref{sec:results_snn}, Table~\ref{tab:prompt_model_configs}).
    \item The multi-agent framework improves the baseline models: on the spiking neural network (SNN) tasks (Tables~\ref{tab:prompt_model_configs} and~\ref{tab:digits_results}), accuracy increases by 13\% on MNIST and by 6\% on the scikit-learn digits; on the physics-informed neural network (PINN) tasks (Section~\ref{sec:results_pinn_one}--\ref{sec:results_ks_gaussian}), the relative error decreases approximately 2500-fold for the generalized nonlinear Schr\"odinger equation (GNLSE) problem with a known analytical solution, and for problems without analytical solutions, the conservation error decreases approximately 325-fold in the selected GNLSE soliton--Gaussian run (more than 100-fold on average over repeated initializations) and the mass drift decreases 14-fold for the Kuramoto--Sivashinsky (KS) wave with a Gaussian perturbation.
\end{enumerate}

\section{Related Work}
\label{sec:related}
 
LLM agents can autonomously plan actions, use tools, and accumulate knowledge while solving multi-step tasks~\cite{xi2023rise}. The ReAct (Reasoning + Acting) pattern has become a baseline architecture~\cite{yao2023react}. Recent surveys~\cite{guo2024large,huang2024understanding} identify typical components: a planning module (Chain-of-Thought~\cite{wei2022chain}, Tree-of-Thoughts~\cite{yao2023tree}), dynamic memory and self-reflection~\cite{shinn2023reflexion}, calls to external tools, and multi-agent designs with role separation~\cite{chen2024survey}. When agents act as meta-optimizers of neural network models, three of these components are essential: role separation distinguishes hypothesis formulation from its software implementation; logged memory prevents rejected variants from being proposed again; and a sandboxed execution environment is required to run generated code safely.
 
LLM agents have been applied to a diverse set of tasks. SWE-bench~\cite{jimenez2024swebench} and MLE-Bench~\cite{chan2024mlebench} have become standard benchmarks for programming and automated ML: the medal rate of the first published o1-preview configuration with AIDE was 16.9\,\%~\cite{chan2024mlebench}, while later agents increased this value to 64\,\% of medals on MLE-Bench Lite (MLE-STAR~\cite{nam2026mle}, including 36\,\% gold medals) and to approximately 35\,\% across all difficulty levels (R\&D-Agent on GPT-5)~\cite{yang2025r}. Similar multi-agent designs are used in materials discovery and biochemistry~\cite{takahara2025accelerated,shi2026knowledge,ghafarollahi2024protagents,swanson2025virtual}. Voyager~\cite{wang2023voyager} demonstrates lifelong accumulation of a skill library in the form of executable code, which is close to the idea of logged hypotheses used in this work. Systems for automating the scientific cycle (The~AI~Scientist~\cite{lu2024scientist,yamada2025scientistv2}, Agent~Laboratory~\cite{schmidgall2025agent}) and agent-based robot control~\cite{skorokhodov2025agentcontrol} have been developed in parallel; they show that autonomous research is possible in principle but do not incorporate domain-specific knowledge of PINNs or SNNs.
 
PINN quality depends on the choice of architecture, activations, the balance of loss-function terms, the distribution of collocation points, and the representation of inputs and outputs~\cite{moloshnikov2025solitonmethods}; separate heuristics have been proposed for each component, but no unified automatic procedure has been established. Lang-PINN~\cite{he2025langpinn}, the study closest to the agent-based approach, constructs a PINN from a textual task description but focuses on network construction rather than on selecting its functional blocks for a physical problem.
 
For SNNs, automatic approaches to selecting the architecture and its individual blocks rely on evolutionary algorithms and reinforcement learning~\cite{kim2022neural,na2022autosnn}; formally, neural architecture search (NAS) also includes the generation of individual functional modules within a predefined space of architectural variants. Unlike NAS, the proposed approach uses an LLM, exploits domain knowledge and knowledge of model properties, locally modifies a selected module of an existing network without searching a fixed space of variants, and stores the history of hypotheses in natural language, so that each subsequent iteration can use a substantive analysis of previous iterations rather than only their numerical results.
 
To the authors' knowledge, the open literature contains no work in which LLM agents are used as a general tool for automatically tuning individual functional blocks of neural network models without redesigning the network itself. MLE-STAR~\cite{nam2026mle} is the closest agent system; the similarity is limited to two components: logged experiment memory and a sandboxed execution environment. MLE-STAR applies iterative code modifications with an update only when the metric improves on the outer loop, processes the entire ML script, selects the target block with a separate agent, assigns web search to a separate retriever agent, and involves more than ten agents.

\section{Method: Multi-agent LLM framework}
\label{sec:agent}

The framework optimizes a selected functional block while keeping the rest of the task pipeline fixed. Before a run, the Human Researcher specifies the task, the block that the agents may modify, the evaluation metric, the software contract for the generated code, and a baseline. These settings allow the same framework to be applied to different problems without changing its architecture or execution loop.

\begin{figure}
\centering
\includegraphics[width=0.9\linewidth]{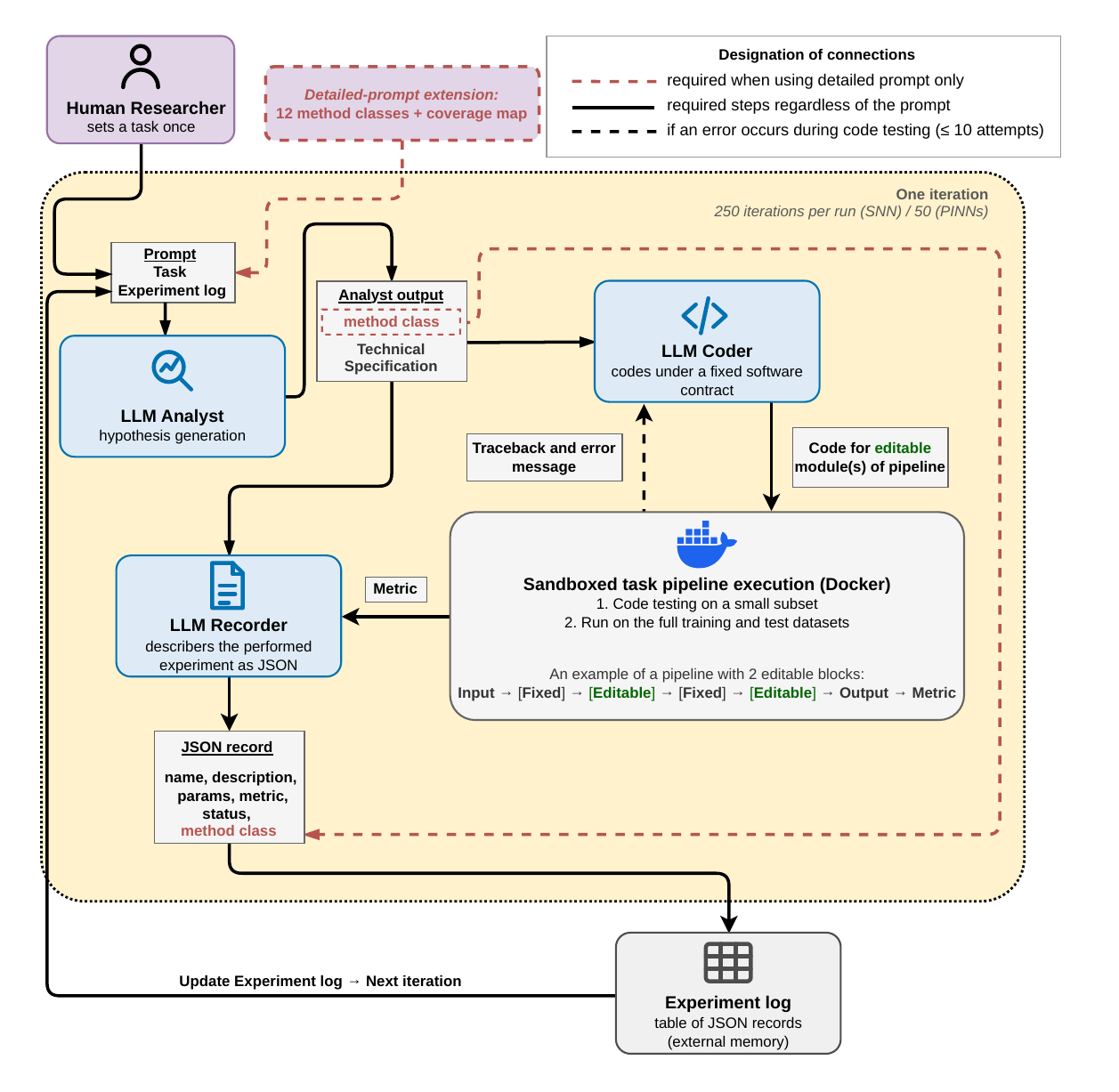}
\caption{Architecture of the multi-agent framework. The Human Researcher configures the task, agent-modifiable block(s), target metric, and software contract once. At each iteration, the LLM Analyst reads the experiment log and produces a technical specification, which the LLM Coder converts into contract-constrained code. The generated module is first functionally tested and then evaluated in a Docker sandbox; runtime errors are returned to the Coder, while a successful evaluation is passed to the LLM Recorder. The Recorder appends a structured JSON record to the experiment log, which serves as the external memory for the next iteration. The detailed-prompt mode additionally provides the Analyst with a method-class taxonomy and a dynamically updated coverage map.
Only task-specific prompts, the software contract, the editable model block, and the target metric change between tasks; the agent roles and loop structure remain unchanged.}
\label{fig:agent_scheme}
\end{figure}

The framework is built on top of an OpenAI-compatible API and separates hypothesis generation, software implementation, and experiment documentation into three LLM roles. A sandboxed execution environment and an external experiment log close the iterative loop (Fig.~\ref{fig:agent_scheme}). The framework consists of five main components:
\begin{itemize}
    \item \textbf{LLM Analyst} receives the task prompt and the current experiment log. It proposes one hypothesis for the agent-modifiable block and expresses the idea as a technical specification with a fixed structure. If the Human Researcher provides a taxonomy of method classes, the Analyst also assigns the hypothesis to one of these classes.

    \item \textbf{LLM Coder} translates the technical specification into Python code. In addition to implementing the proposed method, the code must satisfy a task-specific software contract that defines the allowed dependencies, class names, call signatures, input and output shapes, and trainable parameters.

    \item \textbf{Sandboxed execution environment} (Docker) inserts the generated module into the fixed task pipeline, first performs a short functional test, and then evaluates a valid candidate. A runtime error is returned to the LLM Coder for correction, whereas the metric from a successful run is passed to the logging stage.

    \item \textbf{LLM Recorder} summarizes the specification, validated code, and evaluation result in a structured JSON record, including the method name, short description, and final hyperparameters.

    \item \textbf{Experiment log} is a table of structured records that serves as the external memory of the framework. Each record is assembled from the outputs of the pipeline stages and contains:
    \begin{itemize}
        \item a run identifier assigned by the framework;
        \item the experiment name and a short description generated by the LLM Recorder from the technical specification and validated code;
        \item the execution status assigned by the control loop from the functional-test and evaluation outcomes;
        \item the resulting target-metric value computed by the task-specific evaluation procedure for a successfully evaluated candidate;
        \item the method class selected by the LLM Analyst when a task-specific taxonomy is used;
        \item the final hyperparameter values extracted by the LLM Recorder from the validated generated code.
    \end{itemize}

    Records with the \texttt{With error} status remain stored in the complete experiment log. Before the log is included in the Analyst prompt, the framework filters out these records, so the Analyst receives the baseline and successfully evaluated records only. The run budget counts only successfully evaluated records with the \texttt{OK} status.
\end{itemize}

The technical specification describes the method proposed by the Analyst, whereas the software contract defines the interface and constraints for the generated code. The agents may modify only the selected block; the rest of the task pipeline remains fixed. This separation allows generated code to be inserted and tested automatically. Applying the framework to a new task requires a new prompt, software contract, target metric, and baseline, but the agent roles and execution loop remain unchanged.

\paragraph{Baseline record.}
Each framework run starts with a task-specific baseline record in the experiment log. This record evaluates the fixed pipeline with the simple baseline implementation of the modifiable block and provides the initial reference value of the target metric. The baseline implementations used in this study are defined in Section~\ref{sec:tasks}.

\paragraph{Standard and detailed Analyst prompts.}
The framework supports two modes of hypothesis generation. The \emph{standard prompt} contains the Analyst role, problem statement, target metric, implementation constraints, required format of the technical specification, and complete experiment log.

The \emph{detailed prompt} retains these elements and adds a structured Chain-of-Thought procedure and a list of method classes specified by the Human Researcher for systematic exploration. Each hypothesis is assigned to one principal class, which is also stored in its experiment-log record, and the log is converted into a dynamically updated \emph{coverage map}. For every class, the map reports the number of evaluated hypotheses, the best metric value, and a coverage status such as ``not explored,'' ``limited data,'' or ``explored.'' The map is included in the next Analyst prompt, which prioritizes suitable classes with little or no coverage and revisits an explored class only when a substantially different hypothesis is proposed. This mechanism structures an open natural-language search without reducing it to a finite list of implementations. The task-specific prompt configurations are described in Section~\ref{sec:tasks}; the detailed-mode prompt materials are provided in Supplementary Materials.

\subsection{Run protocol}
\label{sec:protocol}

One framework run is a series with a task-specific number of successful iterations. At each iteration, one hypothesis is tested in six steps:
\begin{enumerate}
    \item The Analyst receives the task prompt and the current experiment log. At the first iteration, the log contains only the baseline record. In detailed mode, the current coverage map is generated from the log and included in the prompt.
    \item The Analyst creates a technical specification according to the prompt template. In detailed mode, the specification also identifies the method class.
    \item The Coder implements the specification subject to the corresponding software contract.
    \item The code undergoes an automatic functional test on a small sample in the sandboxed environment. If the test fails, the error message, generated code, and original specification are returned to the Coder. The Coder analyzes the reported error and rewrites the code while preserving the proposed method and software contract; the revised version is then tested again. If valid code cannot be obtained within 10 attempts, the experiment is assigned the \texttt{With error} status; it does not count toward the run budget, and the framework proceeds to a new hypothesis.
    \item Code that passes the functional test is evaluated according to the task-specific evaluation protocol, and the resulting metric is assigned the \texttt{OK} status. The budgets, metrics, data-use rules, and repetition schemes used in this study are specified in Section~\ref{sec:setup_eval} and the accompanying appendices.
    \item The Recorder combines the specification, validated code, execution result, and fitted hyperparameters into a JSON record and appends it to the log. The updated log becomes the external memory for the next iteration.
\end{enumerate}

\section{Experimental setup}
\label{sec:tasks}

The same framework is applied to five tasks with different types of information representation: two spiking-network classification tasks, two boundary-value problems for the generalized nonlinear Schrödinger equation, and one boundary-value problem for the generalized Kuramoto--Sivashinsky equation.

\subsection{SNN task: Correlation Classwise Network decoder}
\label{sec:setup_snn}
\begin{figure}
\centering
\includegraphics[width=\linewidth]{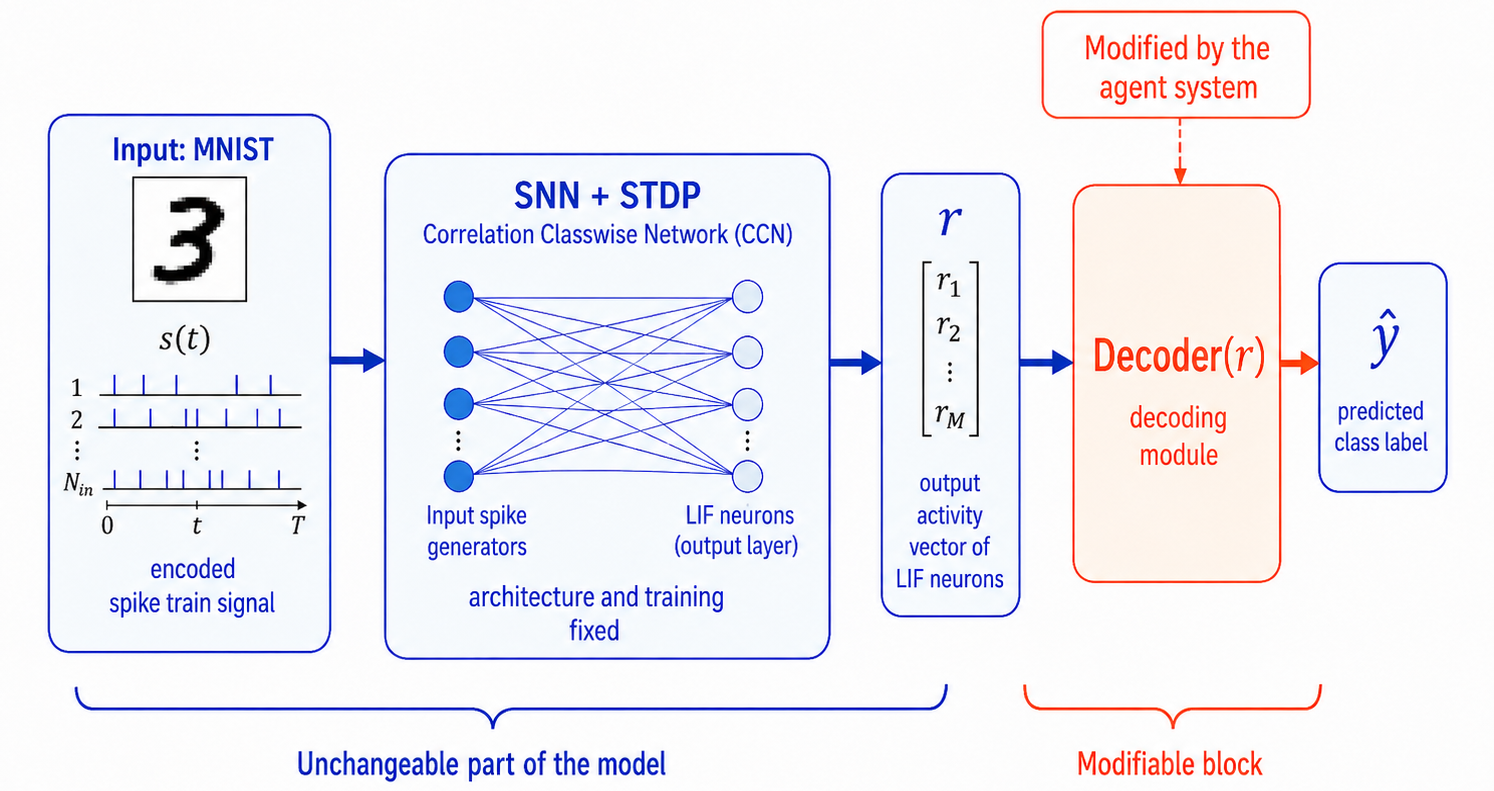}
\caption{SNN pipeline. The input encoding, the SNN trained with STDP, and the output activity vector \(\mathbf{r}\) (left and middle) are fixed. Only the decoder module \(\mathrm{Decoder}(\mathbf{r})\) (right) is generated and optimized by the framework.}
\label{fig:snn_pipeline}
\end{figure}

The classifier is based on a Correlation Classwise Network (CCN) spiking neural network~\cite{sboev2024direct,kunitsyn2025probabilistic}, whose weights are trained using a local spike-timing-dependent plasticity (STDP) rule. Image pixels are normalized and encoded as correlated spike trains that serve as the network input. The output spike activity of the leaky integrate-and-fire (LIF) neurons is then translated into a class label by the decoding module. The \textbf{fixed part}, which the framework cannot modify, comprises the input encoding, the CCN, and the recorded output activity vector \(\mathbf{r}\in\mathbb{R}^{K}\); the \textbf{agent-modifiable part} is the \(\mathrm{Decoder}(\mathbf{r})\) module (Fig.~\ref{fig:snn_pipeline}). Its software contract permits fitting parameters from training activity and class labels but prohibits modifying the trained SNN, spike encoding, or recorded activity and prohibits error backpropagation. The decoder must return one valid class label per sample. We evaluate the method on two datasets: MNIST~\cite{lecun1998} (\(28\times28\) images, \(N=784\) spike generators) and the scikit-learn \texttt{digits} dataset~\cite{scikit-learn} (\(8\times8\), \(N=64\)). Both contain 10 classes, so the network has \(K=10\) output LIF neurons. Accuracy is used as the target metric. As a \textbf{baseline}, we use an \(\arg\max\) decoder: the predicted class corresponds to the output neuron with the highest spike rate. Candidate selection uses only development data; the held-out test sets are unavailable to the agents, as detailed in Section~\ref{sec:snn_protocol}. Full details of the SNN model, the encoding scheme, the STDP rule, and the hyperparameters are given in Appendix~\ref{app:snn}.

\subsection{PINN tasks: GNLSE and Kuramoto--Sivashinsky}
\label{sec:setup_pinn}
Two equations are investigated in this study.
The first is the generalized nonlinear Schrödinger equation with the nonlinear correction~\eqref{eq:gnlse_main} from~\cite{zavertyaev2024opticalsolitons,moloshnikov2025sixthordergnse}, and the second is the generalized Kuramoto--Sivashinsky equation~\eqref{eq:ks_eq} from~\cite{kudryashov2026traveling}.

\begin{equation}
i q_t + q_{xx} + |q|^{2} q\left(1 - \alpha |q|^{2} + \beta |q|^{4}\right) = 0,
\label{eq:gnlse_main}
\end{equation}

where the complex field \(q(x,t)=u(x,t)+iv(x,t)\) is approximated by a fully connected network, while the equation ($\mathcal{L}_{pde}$), initial-condition ($\mathcal{L}_{ic}$), and boundary-condition ($\mathcal{L}_{bc}$) residuals enter the loss functional ($\mathcal{L}$) according to the Physics-Informed Neural Networks framework~\cite{raissi2019physics}, a detailed description is given in the appendix \ref{app:pinn}. In all configurations, \(\beta=0\).

\begin{figure}
    \centering
    \includegraphics[width=0.95\linewidth]{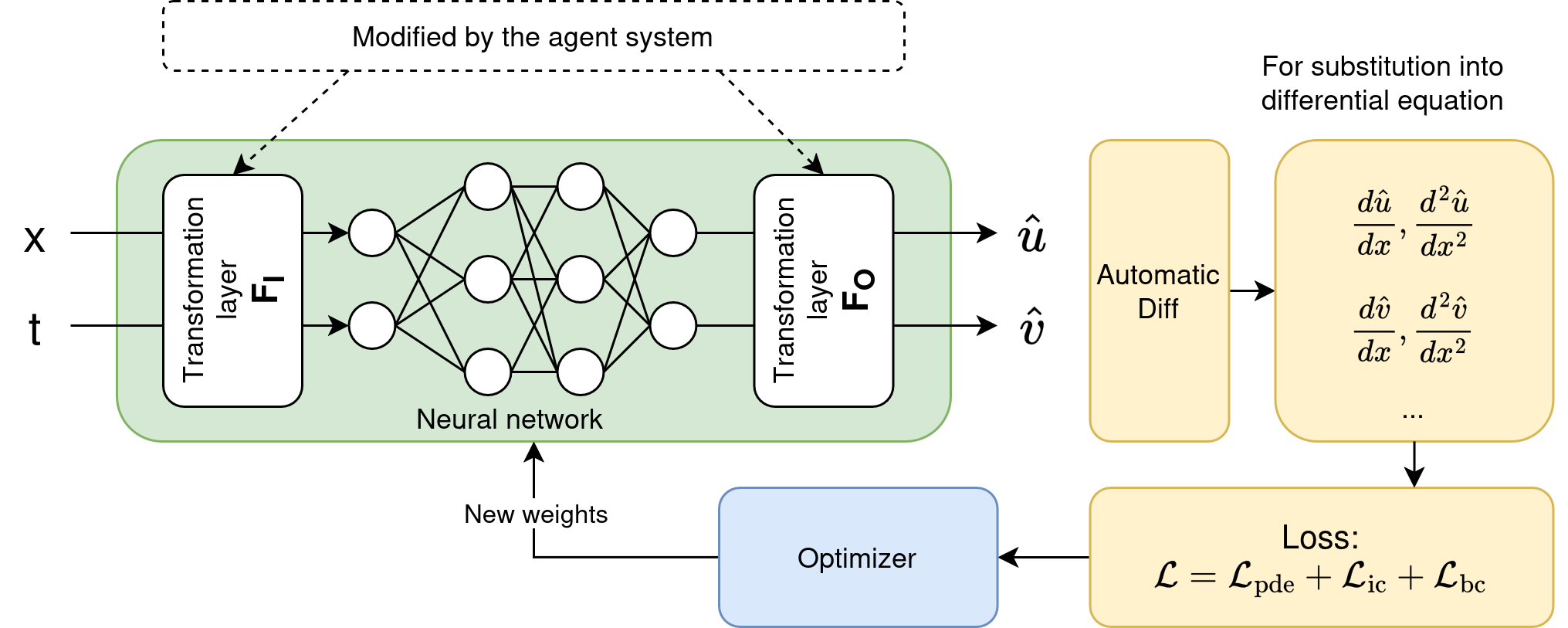}
    \caption{PINN architecture. The coordinates \((x,t)\) pass through the transformation layer \(F_I\), the fully connected network, and the transformation layer \(F_O\), producing predictions \(\hat{u},\hat{v}\). Automatic differentiation computes the derivatives for the loss function.
    The loss function ($\mathcal{L}$) consists of the residual in the differential equation ($\mathcal{L}_{pde}$) and the errors in the initial ($\mathcal{L}_{ic}$) and boundary conditions ($\mathcal{L}_{bc}$).
    The optimizer updates the network weights.}
    \label{fig:pinn_scheme}
\end{figure}

The \textbf{fixed part} comprises the multilayer perceptron (MLP) architecture, loss functional, and training procedure. The \textbf{agent-modifiable part} comprises two transformations (Fig.~\ref{fig:pinn_scheme}): the input transformation \(F_I\), which maps the coordinates \((x,t)\) to the MLP input features, and the output transformation \(F_O\), which maps the MLP outputs and coordinates to the predicted physical field. For the GNLSE problems, the prediction consists of the two real components \((u,v)\); for the Kuramoto--Sivashinsky problem, it is the single real field \(u\). The agents may change the mathematical forms of \(F_I\) and \(F_O\) and may increase the intermediate MLP output dimension when required by \(F_O\), but cannot modify the remaining architecture, loss functional, or training procedure.
In the \textbf{baseline}, \(F_I\) passes \((x,t)\) to the MLP unchanged and \(F_O\) returns the raw MLP outputs.
The framework is applied to two boundary-value problems for~\eqref{eq:gnlse_main}, which differ in the initial condition, domain, and target metric, and to one boundary-value problem for~\eqref{eq:ks_eq}.

\begin{figure}
    \centering
    \includegraphics[width=0.8\linewidth]{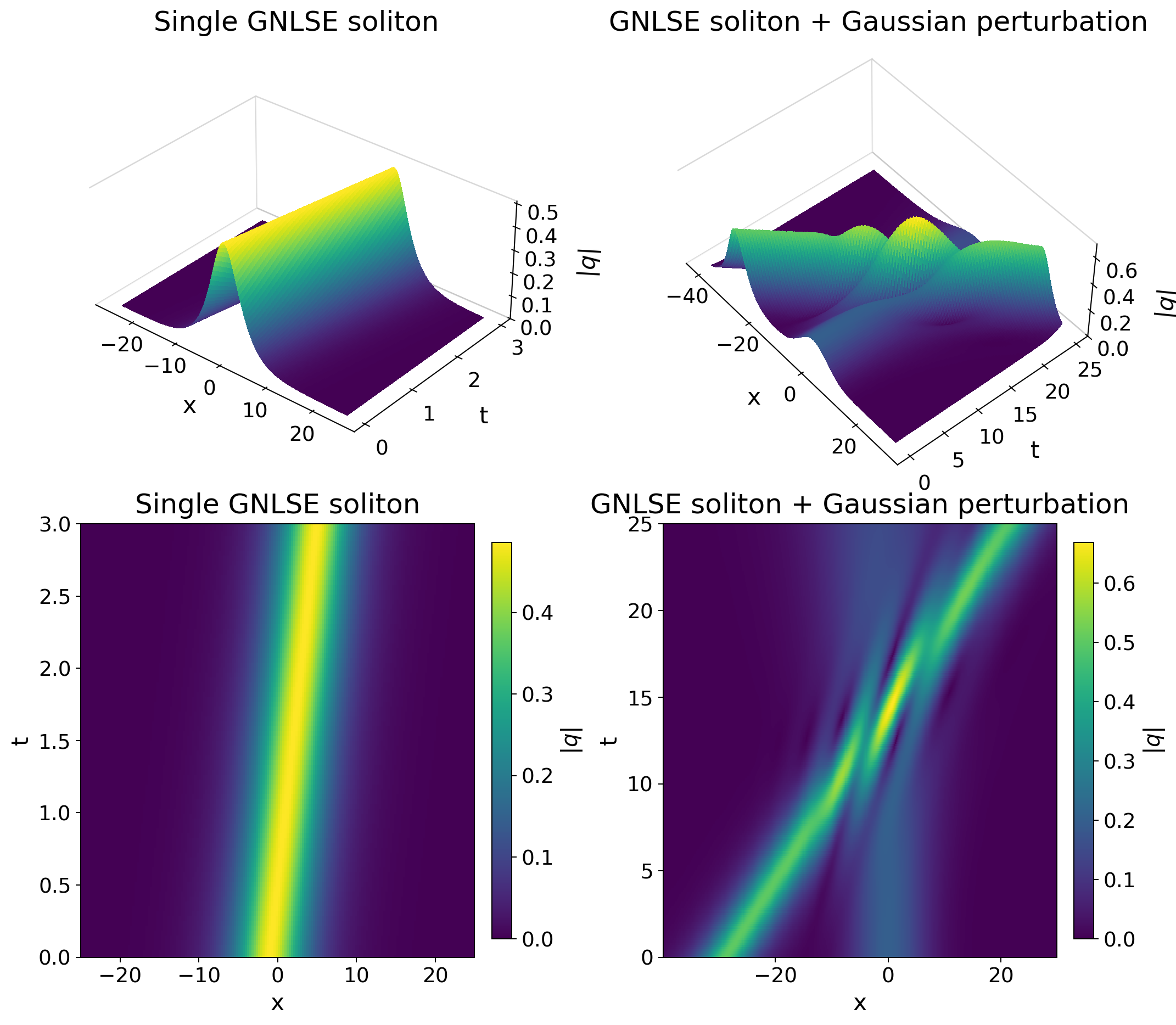}
    \caption{Reference solutions of the two GNLSE boundary-value problems, shown as a surface (top) and a space--time heat map (bottom). Left: the analytical single soliton. Right: the segmented PINN solution of the soliton with a Gaussian perturbation from~\cite{zavertyaev2024opticalsolitons} (Figure~5). The Kuramoto--Sivashinsky problem is not shown because it is solved without a reference solution.}
    \label{fig:pinns_tasks}
\end{figure}

\emph{Single-soliton problem}:
\begin{equation}
\begin{cases}
i\, q_t + q_{xx} + |q|^{2} q\bigl(1 - \alpha|q|^{2} + \beta|q|^{4}\bigr) = 0,\\[2pt]
q(x,\,0) = \tilde{q}(x,\,0),\\[2pt]
q(-25,\,t) = 0, \quad q(25,\,t) = 0,\\[2pt]
(x,t)\in[-25,25]\times[0,3],
\end{cases}
\end{equation}
where \(\alpha=0.1\), \(\beta=0\), and the initial profile \(\tilde{q}\) is specified by the analytical soliton~\eqref{eq:q0} with parameters \(k=1\), \(w=0.88\), \(x_0=0\), \(\theta_0=0\), and \(\mu=4(k^{2}-w)\). Because an analytical solution is known for this problem, quality is evaluated directly using the relative amplitude error \(\mathrm{Rel}_h=\|h_{\mathrm{true}}-h_{\mathrm{pred}}\|_2/\|h_{\mathrm{true}}\|_2\), \(h=\sqrt{u^2+v^2}\). This problem provides a transparent test: the agent can rely on the explicit structure of the solution.

\emph{Soliton problem with a Gaussian perturbation}:
\begin{equation}
\begin{cases}
i\, q_t + q_{xx} + |q|^{2} q\bigl(1 - 0.3|q|^{2}\bigr) = 0,\\[2pt]
q(x,\,0) = q_{2}(x,\,0) + 0.2\,e^{-x^{2}/50},\\[2pt]
q(-50,\,t) = 0, \quad q(50,\,t) = 0,\\[2pt]
(x,t)\in[-50,50]\times[0,25],   
\end{cases}
\end{equation}
where the soliton component \(q_{2}\) has the same form as~\eqref{eq:q0} with \(\alpha=0.3\), \(k=1\), \(w=0.88\), \(x_0=-30\), and \(\theta_0=0\). The Gaussian packet is \(A_g\exp[-(x-B_g)^{2}/(2C_g^{2})]\) with \(A_g=0.2\), \(B_g=0\), and \(C_g=5\).
No analytical solution is known, so the search is driven by two physical metrics.
The mean absolute residual of the boundary-value problem is
\begin{equation}
\mathrm{Res}_{\mathrm{bvp}}
=\tfrac13\Bigl(
\langle|\mathcal{R}_{\mathrm{GNLSE}}|\rangle
+\langle|q_{\mathrm{pred}}(x,0)-q(x,0)|\rangle
+\langle|q_{\mathrm{pred}}|_{\partial}\rangle
\Bigr),
\label{eq:gnlse_bvp_residual}
\end{equation}
where \(\mathcal{R}_{\mathrm{GNLSE}}=iq_t+q_{xx}+|q|^{2}q(1-0.3|q|^{2})\), the averages are taken over a uniform evaluation grid, and \(\partial\) denotes the Dirichlet boundaries \(x=\pm50\). The second primary metric is the relative conservation error of the integrals of motion \(I_1=\int |q|^2\,dx\) and \(I_2=\int(q^*q_x-qq_x^*)\,dx\). For each invariant, we compute its maximum relative drift from the initial value over the considered time interval,
\[
\mathrm{Rel}_{I_j}
=
\frac{\max_t\left|I_j(t)-I_j(0)\right|}
     {\left|I_j(0)\right|},
\qquad j\in\{1,2\}.
\]
These individual errors are combined into the overall conservation metric
\[
\mathrm{Rel}_{I}
=
\mathrm{Rel}_{I_1}+\mathrm{Rel}_{I_2}.
\]

A segmented PINN solution from~\cite{zavertyaev2024opticalsolitons} is retained as an auxiliary amplitude reference \(\mathrm{Rel}_h=\|h_{\mathrm{ref}}-h_{\mathrm{pred}}\|_2/\|h_{\mathrm{ref}}\|_2\). On the agent domain that field itself satisfies \(\mathrm{Res}_{\mathrm{bvp}}=9.11\cdot10^{-5}\) and \(\mathrm{Rel}_i=9.87\cdot10^{-4}\). \(\mathrm{Rel}_h\) has lowest priority: if a hypothesis improves both \(\mathrm{Res}_{\mathrm{bvp}}\) and \(\mathrm{Rel}_i\) relative to this reference, it is regarded as a more accurate solution of the boundary-value problem than the stored PINN field, and \(\mathrm{Rel}_h\) is disregarded.
This problem is the principal substantive test of the method because the agent cannot simply embed a ready-made analytical solution. The complete formulations of both problems, the analytical reference soliton, the collocation-point sampling schemes, and the hyperparameters are provided in Appendix~\ref{app:pinn}.

The second equation considered is a generalization of the Kuramoto--Sivashinsky equation with nonlinear convection~\cite{kudryashov2026traveling}.
This generalization is important in high-temperature plasma research, where the nonlinear dependence of thermal conductivity must be taken into account~\cite{liu1976dissipative,sudan1976stability,chen1977nonlinear}.
We write the source equation for the scalar field \(v(x,t)\) in the normalization
\begin{equation}
v_t + v_{xx} + \sigma v_{xxx} + v_{xxxx}
+ \gamma v v_x + \chi (v^{2})_{xx} = 0.
\label{eq:ks_source}
\end{equation}
Here \(\gamma=2\beta\) is introduced to avoid confusion with the coefficient \(\beta\) in~\eqref{eq:gnlse_main}. For \(\beta=3\sigma\) and \(\sigma=-1\), the coefficients used in this study are \(\gamma=2\beta=6\sigma=-6\) and \(\chi=3\). The corresponding exact traveling wave has the nonzero background \(-1/3\). We therefore remove this background by introducing
\[
v(x,t)=u(x,t)-\frac{1}{3},
\]
so that the neural network approximates the zero-background scalar field \(u(x,t)\). Under this change of variables, Eq.~\eqref{eq:ks_source} becomes the differential equation in~\eqref{eq:ks_eq} (the complete substitution is given in Appendix~\ref{app:ks_shift}).
Unlike~\eqref{eq:gnlse_main}, where the neural network approximates the complex field \(q(x,t)\), this problem consequently has a single real-valued output, and the boundary conditions are zero for both the field and its second derivative.
The general form of the loss functional follows the same three-term PINN scheme as for the GNLSE, but the MLP size, optimizer settings, and training budget differ; they are listed separately in Table~\ref{tab:hyperparams_ks}. The agent-modifiable part comprises the same transformations \(F_I\) and \(F_O\) as in the GNLSE problems.

\emph{Traveling-wave problem with a Gaussian perturbation}:
\begin{equation}
\begin{cases}
u_t-u_{xx}-u_{xxx}+u_{xxxx}
-6u u_x+2u_x+3(u^{2})_{xx}=0,\\[2pt]
u(x,0)
  = \dfrac{1}{2\cosh^{2}(x/2)}
  + 0.02\,\exp\!\left[-\dfrac{(x+6)^{2}}{8}\right],\\[2pt]
u(-30,t)=u_{xx}(-30,t)=0,\quad
u(40,t)=u_{xx}(40,t)=0,\\[2pt]
(x,t)\in[-30,40]\times[0,10].
\end{cases}
\label{eq:ks_eq}
\end{equation}
The unperturbed pulse has the form \(u_{\mathrm{ex}}(x,t)=\frac{1}{2\cosh^{2}((x-t)/2)}\) (speed~\(1\), amplitude~\(1/2\), starting at \(x_0=0\)); the Gaussian packet is located to the left (\(x_g=-6\), width~\(w=2\)) and catches up with the pulse at approximately \(t\approx 6\), after which the perturbation intensifies. The complete analytical solution for this problem is unknown, so quality is assessed only by how well the network satisfies the boundary-value problem itself and by a physical diagnostic. The primary metric of the search is the mean absolute residual of the boundary-value problem,
\begin{equation}
\mathrm{Res}_{\mathrm{bvp}}
=\tfrac14\Bigl(
\langle|\mathcal{R}_{\mathrm{KS}}|\rangle
+\langle|u_{\mathrm{pred}}(x,0)-u(x,0)|\rangle
+\langle|u_{\mathrm{pred}}|_{\partial}\rangle
+\langle|\partial_{xx}u_{\mathrm{pred}}|_{\partial}\rangle
\Bigr),
\label{eq:ks_bvp_residual}
\end{equation}
where \(\mathcal{R}_{\mathrm{KS}}\) is the residual~\eqref{eq:ks_pinn_residual}, the averages are taken over a uniform \(1000\times10\) grid in the domain, and \(\partial\) denotes the two zero boundaries.
The physical diagnostic is the relative mass drift of the predicted solution:
\begin{equation}
\Delta_m^{(u)}
=\max_{k}\frac{\bigl|M_{\mathrm{pred}}(t_k)-M_{\mathrm{pred}}(0)\bigr|}
{\bigl|M_{\mathrm{pred}}(0)\bigr|},
\qquad
M_{\mathrm{pred}}(t)=\int u_{\mathrm{pred}}(x,t)\,dx,
\label{eq:ks_mass_drift}
\end{equation}
with the maximum taken over ten uniformly spaced time slices \(t_k\in[0,10]\).

\subsection{Task-specific prompt configurations}
\label{sec:setup_prompts}

The two universal Analyst modes defined in Section~\ref{sec:agent} are instantiated differently across the experiments. The PINN tasks use only the standard mode, with prompts that specify the differential equation, the modifiable transformations, the software contract, and the physical target metric; they do not use a method-family catalog, Chain-of-Thought procedure, or coverage map.

The SNN classification experiments compare the standard mode, denoted the simple prompt in the result tables, with the detailed mode. The simple prompt specifies the Analyst role, classification problem, target accuracy, data-use and implementation constraints, examples of admissible decoder types, required technical-specification format, and complete experiment log. The detailed prompt adds a researcher-defined taxonomy of 12 decoder-method classes, a structured Chain-of-Thought procedure for coverage and failure analysis, an explicit class field in the technical specification, and the dynamically updated coverage map. The 12 classes define broad areas of exploration rather than 12 fixed algorithms; each class may contain multiple feature transformations, decision rules, and hyperparameter settings, and hypotheses may combine methods from different classes. The complete class taxonomy and prompt texts are provided in Supplementary Section~\ref{app:prompts}.

\subsection{Search baselines and budget: SNN decoder on MNIST}
\label{sec:setup_search_baselines_snn}

The quality reached by the framework depends both on the search performed by the LLM Analyst and on the constraints that the Human Researcher writes into the prompt. To separate these two contributions, we compare the framework with baselines that search the same space of decoders without an Analyst. All comparisons in this subsection are performed on MNIST, where the argmax baseline gives $79.88\,\%$.

\paragraph{Shared search space.}
The decoder catalog is a fixed discrete space of estimators with grids of their hyperparameters, together with the operators that build ensembles and multistage pipelines. It is specified once as a TPOT \texttt{config\_dict} that is listed in Supplementary Section~\ref{app:random_catalog} and its classes follow the decoder taxonomy of the detailed Analyst prompt (Section~\ref{sec:setup_prompts}, Supplementary Materials). The Analyst prompt permits exactly the decoders contained in the catalog. The restriction reaches the systems in different ways: the agent receives it as text and can in principle depart from it, whereas the baselines receive it as a constraint on the sampler. We inspected the Experiment logs of all agent runs and found no record outside the catalog. The space of admissible decoders is therefore common to all systems, and they differ in the search strategy and in the selection protocol described below.

\paragraph{Compared configurations.}
Three baseline configurations are compared with the framework, each with a budget of 250 evaluated decoders per run.

\begin{itemize}
    \item Random Choice is an LLM-based sampler that draws decoder configurations from the same catalog through independent calls to Qwen3-30B-A3B-Instruct-2507 deployed on a rented GPU, without a Chain-of-Thought stage and without a debugging cycle. Each call receives a compressed description of the catalog and returns one decoder configuration in JSON, sampled independently of the previous choices. The configuration is implemented and evaluated by the same sandboxed pipeline as an agent hypothesis. A call that yields a configuration outside the catalog, a configuration that fails the functional test, or a repetition of a configuration already recorded is discarded and does not enter the Experiment log. Calls are issued until the Experiment log contains 250 evaluated decoders.
    \item TPOT~\cite{olson2016tpot} \texttt{classic} is the standard genetic algorithm with a population of 25 individuals and 9 generations, which together with the initial population gives $25\times(9+1)=250$ evaluated candidates.
    \item TPOT \texttt{sequential} uses a population of one individual and 249 consecutive generations, $1\times(249+1)=250$ candidates, so that the budget is preserved while population-level exploration is replaced by a single mutating lineage.
\end{itemize}

In both TPOT modes the search may combine models and build multistage processing sequences, because the corresponding operators are part of the catalog. TPOT samples and mutates pipelines according to its internal grammar, whereas Random Choice draws every configuration independently under an instruction to sample uniformly over the catalog.

\paragraph{Budget.} For the framework, the budget is the number of successfully evaluated records with the \texttt{OK} status. Records with the \texttt{With error} status remain stored in the complete Experiment log but do not count toward the budget and are excluded from the filtered log view passed to the Analyst. For Random Choice, the budget is the number of successfully evaluated decoder records; invalid calls discarded as described above do not count toward it. For TPOT, the budget is the number of fitted pipelines reported by the optimizer. In every configuration the budget is 250 per run, and every configuration was run 10 times. Within one run we take the best accuracy over the 250 evaluated decoders and the position at which it was first reached; Table~\ref{tab:tpot_vs_agent} reports the smallest and the largest of these ten per-run values.

\paragraph{Selection and evaluation.} The baselines were run on the same trained SNN instances and the same output spike matrices as the prompt comparisons of Table~\ref{tab:prompt_model_configs}, so every baseline run is matched with the corresponding agent run and the comparison is paired. The internal scoring rules of the compared systems are not identical. In the framework and in Random Choice, a candidate is scored on a validation set formed from the training set, and the value entered in the Experiment log (for the Analyst) is the mean over three random seeds. The held-out test set is not available to the Analyst (Section~\ref{sec:setup_eval}). TPOT selects its best pipeline by mean accuracy in five-fold cross-validation on the training set, after which the selected pipeline is retrained on the entire training set and evaluated once on the test set. The comparison in Table~\ref{tab:tpot_vs_agent} is therefore made on the test accuracy of the best decoder of a run, which is defined in the same way for all systems. 

\subsection{Evaluation protocol}
\label{sec:setup_eval}

\paragraph{Run budget and thresholds.} The specified run budget (50 for the PINN tasks and 250 for the SNN tasks) counts only successful iterations; see Section~\ref{sec:protocol} for the iteration steps. The target metric threshold (accuracy above 90\,\% for MNIST and 93\,\% for Digits; a decrease in \(\mathrm{Rel}_h\) relative to the baseline for the single-soliton GNLSE problem, a decrease in \(\mathrm{Res}_{\mathrm{bvp}}\) and \(\mathrm{Rel}_i\) for the GNLSE problem with a Gaussian perturbation, and a decrease in \(\mathrm{Res}_{\mathrm{bvp}}\) for the Kuramoto--Sivashinsky problem) provides a reference point for the analyst but is not a stopping criterion. Each hypothesis is evaluated with three random seeds in the single-soliton problem and with a single seed in the remaining two PINN problems, where one training run takes tens of minutes.

\paragraph{Compared configurations.} In both GNLSE problems, the baseline with identity \(F_I,F_O\) transformations is compared with three agent configurations: Perplexity Sonar (\texttt{sonar-pro})~\cite{perplexity2026docs}, Qwen3-30B-A3B-Instruct-2507 (\texttt{qwen-qwen3-30b-a3b-instruct-2507})~\cite{qwen2025qwen3}, and Claude Opus~4.6 (\texttt{anthropic-claude-opus-4.6}); the Analyst prompt is the same for different LLMs.
Because a complete agent run for the Kuramoto--Sivashinsky problem is computationally expensive, we evaluated only Claude Opus~4.6, which produced the best results on the other tasks. The comparison for this problem is therefore limited to the Claude configuration and its baseline; different LLM models are not compared with one another.
The Qwen3-30B-A3B-Instruct-2507 weights~\cite{qwen2025qwen3} were downloaded from the model's public repository. In the classification tasks, pairs \(\langle\)language model, prompt\(\rangle\) are compared: Claude Opus 4.6, Perplexity Sonar, and Qwen3-30B-A3B-Instruct-2507, each with the simple and detailed prompts defined in Section~\ref{sec:setup_prompts}; the baseline is an argmax decoder based on the output spike rates of the LIF neurons. Ablation runs with the detailed prompt and the coverage-map section removed were also performed for all three models, while retaining the method catalog and Chain-of-Thought. Accuracy comparisons are provided in Tables~\ref{tab:qwen_prompt_paired_runs},~\ref{tab:qwen_ablation_prompt_paired_runs},~\ref{tab:sonar_prompt_paired_runs},~\ref{tab:sonar_ablation_prompt_paired_runs},~\ref{tab:opus_prompt_paired_runs}, and~\ref{tab:opus_ablation_prompt_paired_runs}.

\paragraph{Metrics.} In the single-soliton problem, \(\mathrm{Rel}_h\) is the final characteristic of a hypothesis; the best \(\mathrm{Rel}_h\) (mean over three internal seeds) is reported. In the GNLSE problem with a Gaussian perturbation, hypotheses are ranked by the conservation error \(\mathrm{Rel}_i\), while the  boundary-value-problem residual \(\mathrm{Res}_{\mathrm{bvp}}\)~\eqref{eq:gnlse_bvp_residual} is reported alongside it. The robustness of the best configurations was additionally tested using repeated runs (Appendix~\ref{app:pinn_extra}). In the Kuramoto--Sivashinsky problem, hypotheses are ranked by the boundary-value-problem residual \(\mathrm{Res}_{\mathrm{bvp}}\)~\eqref{eq:ks_bvp_residual}, and the mass drift \(\Delta_m^{(u)}\)~\eqref{eq:ks_mass_drift} is reported as an independent physical diagnostic. In the classification tasks, the target metric is test accuracy, but during a run accuracy is evaluated on a validation set formed from the training set so that test-set accuracy is not entered in the experiment log.

\paragraph{Baseline model accuracies.} In the single-soliton problem, the identity-map baselines give \(\mathrm{Rel}_h=4.69\cdot10^{-2}\), \(5.51\cdot10^{-2}\), and \(4.85\cdot10^{-2}\) in the Sonar, Claude, and Qwen agent runs, respectively. In the GNLSE problem with a Gaussian perturbation, the corresponding single-run baselines give \(\mathrm{Rel}_i=1.28\cdot10^{-1}\), \(2.15\cdot10^{-1}\), and \(1.85\cdot10^{-1}\); over \(n=10\) additional baseline trainings, mean is \(\mathrm{Rel}_i=(1.65\pm0.18)\cdot10^{-1}\) (Table~\ref{tab:pinn_repeats}). In the Kuramoto--Sivashinsky problem, the baseline gives \(\mathrm{Res}_{\mathrm{bvp}}=1.41\cdot10^{-3}\) and \(\Delta_m^{(u)}=9.31\cdot10^{-2}\). The SNN baseline (argmax over output spike rates, without Gaussian receptive fields (GRF) or bagging) gives an accuracy of 79.88\,\% on MNIST and 90.83\,\% on \texttt{digits}, consistent with previously published results~\cite{sboev2024direct}. The gap between the datasets motivated target accuracies of 90\,\% on MNIST and 93\,\% on \texttt{digits}.

\subsection{Statistical protocol}
\label{sec:setup_stats}

\paragraph{Repeated MNIST runs and statistical analysis.} To assess the effect of prompt structure, 10 complete runs, each comprising 250 iterations of agent search, were performed for every configuration. Two paired comparisons were performed for each of the three models: the simple prompt was compared with the full detailed prompt, and the detailed prompt with a coverage map was compared with its ablated version without the map. Each comparison comprised ten matched pairs of complete agent runs. Within each pair, the same trained SNN instance and identical output spike data were used, so differences between configurations were evaluated from paired differences in the maximum accuracy achieved.

For Perplexity Sonar, the difference between prompt variants was consistent across all ten runs: with the simple prompt, the maximum accuracy remained at the baseline level, whereas with the detailed prompt it exceeded 90\,\% in every case. For Qwen3-30B-A3B-Instruct-2507, the detailed prompt also increased the maxima relative to the baseline in all ten runs, although the magnitude of the improvement varied considerably. A paired statistical analysis over ten matched runs was performed for all three analyst models.

The prompt effect was tested using a two-sided exact Wilcoxon signed-rank test; two tests were performed for each of the three analyst models---``detailed versus simple'' and ``detailed versus detailed without the coverage map''---giving six tests to which the Holm correction for multiple comparisons was applied. Pairs were formed from runs with the same seed, which determines the framework initialization and the order of data presentation; in the Supplementary tables, the rows are ordered by decreasing accuracy of the detailed prompt rather than by seed number. For configurations in which the accuracy of the simple prompt exactly matched the baseline in every run, the paired test degenerates into a sign test, as reflected in the reported \(W\) and \(p\) values. The paired-analysis methodology and rationale for selecting the test are provided in Supplementary Section~\ref{app:snn_extra}. The unadjusted and Holm-adjusted $p$ values, together with the $W$ statistic for all six tests, are reported in Supplementary Section~\ref{app:snn_extra}; the difference is statistically significant in all six tests at $\alpha=0.05$ after the Holm correction.

\paragraph{Repeated GNLSE trainings.} For the soliton with a Gaussian perturbation, the identity baseline and the two Claude configurations selected by \(\mathrm{Res}_{\mathrm{bvp}}\) and \(\mathrm{Rel}_i\) were each retrained ten times from independent random initializations. The random seeds were not recorded, so the three series are unpaired. The exported repeated-run data contain \(\mathrm{Rel}_i\) and its two components but not \(\mathrm{Res}_{\mathrm{bvp}}\); consequently, this experiment characterizes robustness of invariant preservation only and does not establish a repeated-run success rate for the joint criterion.

\section{Results}
\label{sec:results}

\subsection{Summary across five tasks}
\label{sec:results_summary}

The final results obtained by applying the same agent framework to five tasks are summarized in Table~\ref{tab:summary}. Accuracy (higher is better) is used for the classification tasks, whereas a relative error or physical diagnostic (lower is better) is used for the PINN tasks. The improvement over the baseline is expressed in task-specific units: for classification tasks, the absolute accuracy increase \(\mathrm{Acc}-\mathrm{Acc}_{\mathrm{base}}\) is reported in percentage points, whereas for PINN tasks the error-reduction factor \(\mathrm{Err}_{\mathrm{base}}/\mathrm{Err}\) is reported. The accuracy values and improvements shown in the table are rounded to integers. The corresponding unrounded accuracies are 79.88 and 92.83\,\% for MNIST (Table~\ref{tab:prompt_model_configs}) and 90.83 and 96.67\,\% for \texttt{digits} (Table~\ref{tab:digits_results}).

For the GNLSE soliton with a Gaussian perturbation, Table~\ref{tab:summary} reports the conservation error \(\mathrm{Rel}_i\). For the Kuramoto--Sivashinsky problem, it reports the mass drift \(\Delta_m^{(u)}\), an independent physical diagnostic recorded alongside the optimized boundary-value-problem residual \(\mathrm{Res}_{\mathrm{bvp}}\); the latter decreases by a factor of 9.3 in the same run (Table~\ref{tab:ks_gauss_results}). Each row compares a single best agent result with the corresponding identity baseline from the same analyst run.

For the GNLSE soliton with a Gaussian perturbation, robustness was additionally evaluated over ten independent trainings. These runs give \(\mathrm{Rel}_i=(1.65\pm0.18)\cdot10^{-1}\) for the baseline, \((1.42\pm0.59)\cdot10^{-3}\) for the configuration selected by \(\mathrm{Res}_{\mathrm{bvp}}\), and \((1.60\pm0.81)\cdot10^{-3}\) for the configuration selected by \(\mathrm{Rel}_i\). The corresponding reductions relative to the baseline mean are \(116\times\) and \(103\times\), respectively (Table~\ref{tab:pinn_repeats}). Overall, the best agent configuration outperforms the baseline on the reported metric in all five tasks.

\begin{table}[htbp]
\caption{Baseline and best agent results across five tasks.}
\label{tab:summary}
\begin{tabular*}{\tblwidth}{
  @{\extracolsep{\fill}}lccc@{}
}
\toprule
Task (metric) &
\makecell{Baseline\\(single run)} &
\makecell{Best agent result\\(best in series)} &
\makecell{Improvement over baseline\\(pp / factor)} \\
\midrule
SNN, MNIST (Acc) &
80\,\% &
93\,\% &
\(+13\)~pp \\
SNN, scikit-learn digits (Acc) &
91\,\% &
97\,\% &
\(+6\)~pp \\
\midrule
PINN, GNLSE, single soliton (\(\mathrm{Rel}_h\)) &
\(5.51\cdot10^{-2}\) &
\(2.21\cdot10^{-5}\) &
\(2491\times\) \\
PINN, GNLSE, soliton + Gaussian (\(\mathrm{Rel}_i\))
& \(2.15\cdot10^{-1}\)
& \(6.61\cdot10^{-4}\)
& \(325\times\) \\
PINN, KS, traveling wave + Gaussian (\(\Delta_m^{(u)}\))
& \(9.31\cdot10^{-2}\)
& \(6.56\cdot10^{-3}\)
& \(14\times\) \\
\bottomrule
\end{tabular*}
\end{table}

\subsection{SNN classification results}
\label{sec:results_snn}

The agent configurations on MNIST are compared in Table~\ref{tab:prompt_model_configs}. For Claude Opus~4.6, the detailed prompt raises the mean maximum accuracy over ten runs paired by seed from $91.13\pm0.44\,\%$ to $92.15\pm0.62\,\%$, and the best result from $91.65\,\%$ to $92.83\,\%$. It gives the higher run maximum in eight of the ten pairs, and the two-sided exact Wilcoxon signed-rank test yields $W=5$, $p=0.0195$, which remains significant after the Holm correction over the six tests of Section~\ref{sec:setup_stats}. The paired gain averages $1.02\pm0.90$~percentage points (Table~\ref{tab:opus_prompt_paired_runs}), so for this model the effect is comparable to the spread between runs. For Perplexity Sonar and Qwen3-30B-A3B-Instruct-2507, none of the 250 decoders proposed per run under the simple prompt outperformed the baseline, so the run maximum for both models matched the baseline of $79.88\,\%$ in all ten runs (std \(=0.00\)). The detailed prompt lifted both models above the baseline, though to different extents: for Perplexity Sonar, the individual-run maxima ranged from $90.93\,\%$ to $92.70\,\%$ with a mean of $91.76\pm0.56\,\%$, so the $90\,\%$ reference point was exceeded in all ten runs, whereas for Qwen3-30B-A3B-Instruct-2507 the mean maximum was $85.88\pm2.42\,\%$ and the best run reached $89.16\,\%$ without crossing it. For both models the prompt effect is significant by the same paired test ($W=0$, $p=0.00195$, the minimum attainable value at $n=10$; Tables~\ref{tab:sonar_prompt_paired_runs} and~\ref{tab:qwen_prompt_paired_runs}).

On the scikit-learn digits dataset, the effect of the detailed prompt depended on the analyst model. For Claude Opus 4.6, the mean run maximum increased from \(95.07\pm0.52\,\%\) to \(96.12\pm0.66\,\%\), with all ten runs reaching 93\,\%. For Perplexity Sonar, the mean changed from \(93.44\pm0.57\,\%\) to \(93.31\pm0.93\,\%\), while the number of runs reaching 93\,\% decreased from 9/10 to 8/10, although the best observed result increased from 93.89\,\% to 94.17\,\%. For Qwen3-30B-A3B-Instruct-2507, the mean increased from \(91.63\pm0.74\,\%\) to \(93.66\pm0.79\,\%\), and the number of runs reaching 93\,\% increased from 0/10 to 7/10.

\begin{table}[htbp]
\centering
\caption{Comparison of analyst models and prompt variants on MNIST. For each configuration, the table reports the number of complete agent runs, the mean and standard deviation of the maximum accuracy achieved within an individual run, the highest accuracy across all runs, and the number of runs in which the 90\,\% threshold was reached. Ten matched runs were performed for every configuration.}
\label{tab:prompt_model_configs}
\resizebox{\linewidth}{!}{%
\begin{tabular}{@{\extracolsep{\fill}}llccc@{}}
\toprule
Model &
Prompt &
\makecell{Mean maximum\\accuracy \(\pm\) std, \%} &
\makecell{Best\\accuracy, \%} &
\makecell{Runs with\\accuracy \(\geq90\,\%\)} \\
\midrule
Claude Opus~4.6
& simple
& \(91.13\pm0.44\)
& 91.65
& 10 out of 10 \\
Claude Opus~4.6
& detailed
& \(92.15\pm0.62\)
& \textbf{92.83}
& 10 out of 10 \\
Claude Opus~4.6
& \makecell[l]{detailed \\without coverage map}
& \(91.02\pm0.25\)
& 91.45
& 10 out of 10 \\
\midrule
Perplexity Sonar
& simple
& \(79.88\pm0.00\)
& 79.88
& 0 out of 10 \\
Perplexity Sonar
& detailed
& \(91.76\pm0.56\)
& 92.70
& 10 out of 10 \\
Perplexity Sonar
& \makecell[l]{detailed \\without coverage map}
& \(82.34\pm0.70\)
& 83.18
& 0 out of 10 \\
\midrule
Qwen3-30B-A3B-Instruct-2507
& simple
& \(79.88\pm0.00\)
& 79.88
& 0 out of 10 \\
Qwen3-30B-A3B-Instruct-2507
& detailed
& \(85.88\pm2.42\)
& 89.16
& 0 out of 10 \\
Qwen3-30B-A3B-Instruct-2507
& \makecell[l]{detailed \\without coverage map}
& \(80.18\pm0.45\)
& 81.15
& 0 out of 10 \\
\bottomrule
\end{tabular}%
}
\end{table}

\begin{table}[htbp]
\centering
\small
\setlength{\tabcolsep}{4pt}
\renewcommand{\arraystretch}{1.2}
\caption{Comparison of the agent system, TPOT evolutionary search
(\texttt{classic} and \texttt{sequential} modes), and the Random Choice
baseline for the SNN decoder on MNIST with a budget of 250 evaluated models.}
\label{tab:tpot_vs_agent}
\begin{tabular}{lccccc}
\toprule
\makecell{Method} &
\makecell{Mode} &
\makecell{\(\mathrm{Acc}_{min}\), \%} &
\makecell{\(\mathrm{Acc}_{max}\), \%} &
\makecell{\(\mathrm{k}_{min}\)} &
\makecell{\(\mathrm{k}_{max}\)} \\
\midrule
TPOT & classic    & 89.60 & 90.15 & 201 & 243 \\
TPOT & sequential & 89.38 & 89.93 & 67 & 210 \\
Random Choice & LLM sampler & 89.48 & 89.93 & - & - \\
\midrule
\makecell[l]{Qwen3-30B-\\A3B-Instruct-2507} & detailed
  & 82.28 & 89.16 & 15 & 209 \\
Perplexity Sonar & detailed
  & 90.93 & 92.70 & 8 & 147 \\
\textbf{Claude Opus 4.6} & \textbf{detailed}
  & \textbf{91.22} & \textbf{92.83} & \textbf{2} & \textbf{43} \\
\bottomrule
\end{tabular}

\begin{flushleft}
\footnotesize
Every \texttt{Method} was run 10 times. Within one run we take the best accuracy over the 250 evaluated models. The minimum and maximum reported values were derived from these 10 values:
\begin{itemize}
\item \(\mathrm{Acc}_{min}\), \(\mathrm{Acc}_{max}\)~--- the smallest and the largest best accuracies.

\item \(\mathrm{k}_{min}\), \(\mathrm{k}_{max}\)~--- the smallest and the largest positions of evaluated models at which the best accuracy was first reached within a single run.
\end{itemize}

Random Choice issues independent calls, so the position at which the run maximum appears does not characterise search progress and $k_{\min}$ and $k_{\max}$ are not reported.
\end{flushleft}
\end{table}

\subsection{Cost of the agent search}
\label{subsec:cost}

The positions \(k_{\min}\) and \(k_{\max}\) in Table~\ref{tab:tpot_vs_agent} characterize the length of the search, not its computational cost. The framework and the Random Choice baseline are therefore compared by three quantities: the token usage per run, the cost of a complete run \(C_{\mathrm{run}}\), and the cost \(C^{*}\) accumulated up to the earliest position \(k_{\min}\). The token usage depends on the structure of the Analyst input, on the number of Coder attempts per iteration (Section~\ref{sec:protocol}), and on the size of the Recorder input.

\paragraph{Rates.} The Coder and the Recorder use gpt-5-mini in all configurations, at the provider's public rate \$0.25/\$2.00 per M input/output tokens. For the Analyst, the rates in effect in June 2026 were \$5/\$25 per M input/output tokens for Claude Opus~4.6 and \$1/\$1 per M for Perplexity Sonar. Qwen3-30B-A3B-Instruct-2507 is deployed on a rented GPU and has no provider price, so its cost and the cost of Random Choice are given at the managed hosting rate for the same model class, \$0.15/\$0.60 per M tokens. Random Choice uses the same Qwen model in a single-agent configuration with 250 evaluated decoders without a Chain-of-Thought stage and without a debugging cycle. Its token usage was measured from the run logs in the same way as for the agent configurations.

\begin{table}[htbp]
\centering
\small
\setlength{\tabcolsep}{4pt}
\caption{Performance, resources, and cost of the agent framework and the Random Choice baseline on MNIST (detailed prompt, budget of 250 evaluated decoders per run). The performance block is taken from Table~\ref{tab:tpot_vs_agent}.}
\label{tab:cost_summary_mnist}
\begin{tabular}{lcccc}
\toprule
Metric & Claude det. & Sonar det. & Qwen det. & \makecell{Random\\Choice} \\
\midrule
\multicolumn{5}{l}{\emph{Performance}} \\
\(\mathrm{Acc}_{\min}\), \%  & $91.22$ & $90.93$ & $82.28$ & $89.48$ \\
\(\mathrm{Acc}_{\max}\), \%  & $92.83$ & $92.70$ & $89.16$ & $89.93$ \\
\(k_{\min}\)                 & $2$     & $8$     & $15$    & --- \\
\(k_{\max}\)                 & $43$    & $147$   & $209$   & --- \\
\midrule
\multicolumn{5}{l}{\emph{Resources per run, tokens in millions (M)}} \\
Evaluated decoders           & $250$   & $250$   & $250$   & $250$ \\
Analyst calls, $N_{\mathrm{an}}$ & $265$   & $252$   & $255$   & $347$ \\
Programmer calls / iter           & $3.31$  & $1.70$  & $2.05$  & --- \\
Analyst in / out, M          & $11.10$ / $1.17$ & $10.89$ / $0.58$ & $11.20$ / $0.58$ & $0.28$ / $0.042$ \\
Programmer in / out, M            & $12.98$ / $4.66$ & $2.40$ / $1.73$  & $2.79$ / $1.32$  & --- \\
Recorder in / out, M         & $2.61$ / $0.058$ & $1.71$ / $0.061$ & $1.38$ / $0.063$ & --- \\
\midrule
\multicolumn{5}{l}{\emph{Cost}} \\
Analyst rate, \$/M in\,/\,out    & $5$ / $25$ & $1$ / $1$ & $0.15$ / $0.60$ & $0.15$ / $0.60$ \\
Programmer, Recorder rate, \$/M       & \multicolumn{3}{c}{{$0.25$ / $2.00$ (gpt-5-mini)}} & --- \\
$C_{\mathrm{an}}$, \$        & $84.66$ & $11.47$ & $2.03$  & $0.07$ \\
$C_{\mathrm{pr}}$, \$        & $12.56$ & $4.06$  & $3.35$  & --- \\
$C_{\mathrm{rec}}$, \$       & $0.77$  & $0.55$  & $0.47$  & --- \\
$C_{\mathrm{run}}$, \$       & $97.99$ & $16.08$ & $5.85$  & $0.07$ \\
$C^{*}$ up to $k_{\min}$, \$ & $0.37$  & $0.21$  & $0.26$  & $0.022$ \\
\bottomrule
\end{tabular}
\end{table}

\paragraph{Token usage.}
The cost of a run is the sum of the token costs of the three roles of the framework, the Analyst, the Programmer, and the Recorder. For a role \(r\),

\begin{equation}
C_{r} = p_r^{\mathrm{in}}\,T_{r}^{\mathrm{in}} + p_r^{\mathrm{out}}\,T_{r}^{\mathrm{out}},
\qquad
C_{\mathrm{run}} = C_{\mathrm{an}} + C_{\mathrm{pr}} + C_{\mathrm{rec}},
\label{eq:cost}
\end{equation}

where \(p_r\) are the per-token rates and \(T_r\) are the token counts taken from the run logs.

The Analyst input at iteration \(i\) consists of the fixed part of the detailed prompt (the role, problem statement, constraints, catalog of 12 method classes, coverage map, Chain-of-Thought instruction, and decoder examples), \(F=4085\) tokens, and the experiment log of \(i\) records of \(d_{\log}\) tokens each. A log record is passed to the Analyst as a JSON object with the fields \texttt{run\_id}, \texttt{name}, \texttt{description}, \texttt{accuracy}, and \texttt{method\_class}, so its footprint \(d_{\log}\) in the prompt exceeds the length \(d\) of the description text itself, which is what the Recorder returns. Over the \(N=250\) accepted iterations of a run the Analyst also makes \(N_{\mathrm{an}}-N\) additional calls that failed to produce an accepted decoder; at each such call the Analyst reads the log at its current length. Approximating the mean log length at the moment of a failed call by \(d_{\log}N/2\)~--- the average log length over the run~--- the total Analyst input is

\begin{equation}
T_{\mathrm{an}}^{\mathrm{in}} = N_{\mathrm{an}} F + d_{\log}\left(\frac{N(N+1)}{2} + (N_{\mathrm{an}} - N)\,\frac{N}{2}\right).
\label{eq:analyst_tokens}
\end{equation}

The first term is the fixed prompt part paid at each of the \(N_{\mathrm{an}}\) Analyst calls; the second term collects the log tokens read during the \(N\) accepted iterations (exact sum) and during the \(N_{\mathrm{an}}-N\) failed calls (approximate).

The mean length of a log record in the Analyst prompt, measured with the \texttt{cl100k\_base} tokenizer as the increment of the prompt length caused by one accepted record, is \(d_{\log}=301\) tokens for Claude Opus~4.6, 312 for Perplexity Sonar, and 317 for Qwen3-30B-A3B-Instruct-2507; the description text alone accounts for 231, 244, and 252 of them. The numbers of Analyst calls per accepted iteration, measured on the same runs, gives \(N_{\mathrm{an}}=265\), 252, and 255 (Table~\ref{tab:cost_summary_mnist}). Equation~\eqref{eq:analyst_tokens} with these values gives \(T_{\mathrm{an}}^{\mathrm{in}}=11.10\), 10.89, and 11.20~M input tokens per run, of which the fixed-prompt contribution \(N_{\mathrm{an}}F\) is 1.08, 1.03, and 1.04~M and the log contribution is 10.02, 9.86, and 10.16~M. The Analyst output contains the Chain-of-Thought reasoning and the technical specification of the method. Its mean length per call, measured on the same runs, is 4399 tokens for Claude Opus~4.6, 2315 for Perplexity Sonar, and 2291 for Qwen3-30B-A3B-Instruct-2507, so that \(T_{\mathrm{an}}^{\mathrm{out}}=N_{\mathrm{an}}\cdot T_{\mathrm{an,call}}^{\mathrm{out}}=1.17\), 0.58, and 0.58~M per run.

The Coder receives its fixed prompt of 612 tokens together with the current specification and returns the Python implementation of the \texttt{Decoding} class; the mean length of accepted code is 4982 tokens for Claude Opus~4.6, 3749 for Perplexity Sonar, and 2513 for Qwen3-30B-A3B-Instruct-2507. If the code does not pass the functional test, the failing code and its trace stay in the dialogue and are re-read at every subsequent attempt of the same iteration, so the Coder input grows quadratically with the length of the debugging chain. The debug rate, measured from the runs, is 3.31 Coder calls per accepted iteration for Claude Opus~4.6, 1.70 for Perplexity Sonar, and 2.05 for Qwen3-30B-A3B-Instruct-2507, with mean failed-attempt lengths of 5991, 4641, and 2724 tokens, respectively. Reconstructing every Coder dialogue of the runs token by token gives Coder token counts per run of 12.98, 2.40, and 2.79~M for the input and 4.66, 1.73, and 1.32~M for the output.

The Recorder receives its own prompt of 99 tokens together with the Coder prompt and the accepted code, and returns the description of length \(d\), so that \(T_{\mathrm{rec}}^{\mathrm{in}}=2.61\), 1.71, and 1.38~M and \(T_{\mathrm{rec}}^{\mathrm{out}}=N d=0.058\), 0.061, and 0.063~M per run. Summation of~\eqref{eq:cost} over the three roles gives complete-run costs of \$97.99, \$16.08, and \$5.85 for Claude Opus~4.6, Perplexity Sonar, and Qwen3-30B-A3B-Instruct-2507, with the Analyst contributing \$84.66, \$11.47, and \$2.03, and costs up to \(k_{\min}\) of \$0.37, \$0.21, and \$0.26.

\paragraph{Cost of Random Choice.}
Random Choice gives the lowest token usage among the compared systems. Each mini-call contains the catalog of method classes as input and one JSON decoder configuration as output; at about 800 input and 120 output tokens per call and 347 calls per run, the run requires 0.28~M input and 0.042~M output tokens. These values are 40 and 14 times smaller than the Analyst token counts of the Qwen agent with the detailed prompt (11.20~M and 0.58~M). In accuracy the two configurations are close: the per-run maxima of Random Choice lie between 89.48 and 89.93\,\%, those of the Qwen agent between 82.28 and 89.16\,\% (Table~\ref{tab:tpot_vs_agent}), and neither configuration reaches the 90\,\% threshold. When Qwen3-30B-A3B-Instruct-2507 is used free of charge, \(C_{\mathrm{run}}=0\) for both configurations, and they differ only in token usage and in \(C^{*}\). At the managed rate for this model class, the Analyst cost differs by a factor of 29 (\$0.07 for Random Choice against \$2.03 for the agent framework with Qwen as the LLM Analyst), and with the Coder and the Recorder included by a factor of 84 (\$0.07 against \$5.85, Table~\ref{tab:cost_summary_mnist}). Sampling of the same catalog using Random Choice method is therefore two orders of magnitude cheaper than the Qwen agent framework at comparable accuracy. Both proprietary Analysts with the detailed prompt exceed it: their minimum run maxima are 90.93\,\% for Perplexity Sonar and 91.22\,\% for Claude Opus~4.6, against 89.93\,\% for Random Choice. The framework is thus preferable to sampling of the same catalog when the Analyst model is sufficiently strong; for a weak Analyst, the additional cost and the additional components of the framework are not compensated by the accuracy.

\subsection{PINN results: single soliton}
\label{sec:results_pinn_one}

The first PINN problem (GNLSE, \(\alpha=0.1\), \(\beta=0\)) has an analytical solution and therefore provides a transparent test of the framework: the agent can embed the explicit form of the one-soliton solution in the \(F_I,F_O\) blocks, and the resulting metrics reflect the correctness of such a ``hinted'' representation rather than an independent search for a nontrivial technique. The results are summarized in Table~\ref{tab:pinn_one_results}. Because each agent run includes its own three-seed identity baseline, each reduction factor is calculated against the baseline from the same run. The best Claude Opus~4.6 hypothesis gives \(\mathrm{Rel}_h=2.21\cdot10^{-5}\), a \(2491\)-fold reduction from \(5.51\cdot10^{-2}\). The best \texttt{sonar-pro} hypothesis reduces \(\mathrm{Rel}_h\) from \(4.69\cdot10^{-2}\) to \(1.97\cdot10^{-4}\) (\(238\times\)), while Qwen3-30B-A3B-Instruct-2507 reduces it from \(4.85\cdot10^{-2}\) to \(8.86\cdot10^{-3}\) (\(5.48\times\)). Because the problem permits substitution of the analytical formula, these figures should not be interpreted as a measure of a complex search; the substantive test is deferred to the second problem.

\begin{table}[H]
\centering
\caption{PINN results for the single soliton (\(\mathrm{Rel}_h\) metric; problem with an analytical solution). Each value is the mean over three internal seeds. Because baseline variation is non-negligible, the change for each analyst is calculated relative to the identity baseline from the same agent run.}
\label{tab:pinn_one_results}
\resizebox{\linewidth}{!}{
\begin{tabular}{@{\extracolsep{\fill}}lccc@{}}
\toprule
Model & Baseline \(\mathrm{Rel}_h\) & Best \(\mathrm{Rel}_h\) & Change relative to baseline \\
\midrule
\texttt{sonar-pro} & \(4.69\cdot 10^{-2}\) & \(1.97\cdot 10^{-4}\) & \(238\times\) better \\
\texttt{anthropic-claude-opus-4.6} & \(5.51\cdot 10^{-2}\) & \(2.21\cdot 10^{-5}\) & \(2491\times\) better \\
\texttt{qwen-qwen3-30b-a3b-instruct-2507} & \(4.85\cdot 10^{-2}\) & \(8.86\cdot 10^{-3}\) & \(5.5\times\) better \\
\bottomrule
\end{tabular}}
\end{table}

The best \texttt{sonar-pro} configuration is \texttt{tw\_nonlinear\_pinn\_core}. Its \(F_I\) supplies a normalized traveling-wave coordinate, normalized time, and a local nonlinearity-density feature; \(F_O\) uses the analytical traveling-wave amplitude and phase and interprets the MLP outputs as bounded amplitude and phase corrections. The best Claude Opus~4.6 configuration, \texttt{PCDS-LCC-EXP-NLOG-MIXCROSS}, uses seven analytical features, including normalized and logarithmic envelopes and two amplitude--time cross-features, while \(F_O\) applies time-gated exponential amplitude and additive phase corrections with separate envelope weights. The best Qwen configuration, \texttt{LogPhaseHybridNorm}, normalizes the analytical soliton coordinate at the input and reconstructs the field from the analytical amplitude and carrier phase multiplied by learned complex corrections. Thus, all three best configurations exploit the known soliton structure in both the representation and the reconstruction. A detailed analysis of the hypotheses is provided in Appendix~\ref{app:pinn_analysis}.
\subsection{PINN results: soliton with gaussian perturbation}
\label{sec:results_pinn_gauss}

The second PINN problem (GNLSE, \(\alpha=0.3\), \(\beta=0\); the initial condition is a soliton with a superimposed Gaussian perturbation) has no analytical solution, so the agent cannot embed a ready-made formula and must search for a substantive representation.
The search is driven by \(\mathrm{Res}_{\mathrm{bvp}}\)~\eqref{eq:gnlse_bvp_residual} and \(\mathrm{Rel}_i\); the reruns reported here do not include \(\mathrm{Rel}_h\).
The human-designed PINN from prior work has \(\mathrm{Res}_{\mathrm{bvp}}=9.11\cdot10^{-5}\) and \(\mathrm{Rel}_i=9.87\cdot10^{-4}\).
The single-run results are summarized in Table~\ref{tab:pinn_gauss_results}. The shared identity baseline combines the lowest \(\mathrm{Rel}_i\) and the lowest \(\mathrm{Res}_{\mathrm{bvp}}\) observed across the analyst runs, and each reduction factor is computed relative to its \(\mathrm{Rel}_i\). The \(325\times\) improvement is obtained relative to the identity baseline from the same Claude Opus~4.6 run, whereas the \(194\times\) improvement is obtained relative to the shared cross-run reference baseline. The best Claude Opus~4.6 configuration by conservation error, \texttt{SPM\_XPM\_CorrWidth\_SplitScale\_HighP}, reduces \(\mathrm{Rel}_i\) from \(1.28\cdot10^{-1}\) to \(6.61\cdot10^{-4}\) (\(194\times\)) while giving \(\mathrm{Res}_{\mathrm{bvp}}=3.95\cdot10^{-5}\). A different Claude configuration, \texttt{SplitAlpha\_ExpPhase\_HighP}, gives the lowest residual, \(\mathrm{Res}_{\mathrm{bvp}}=2.08\cdot10^{-5}\), with \(\mathrm{Rel}_i=1.40\cdot10^{-3}\). The best \texttt{sonar-pro} hypothesis by \(\mathrm{Rel}_i\) improves this metric \(20\times\), but its \(\mathrm{Res}_{\mathrm{bvp}}=2.37\cdot10^{-2}\) is worse than the shared baseline; the best Qwen hypothesis improves \(\mathrm{Rel}_i\) only \(1.36\times\) and likewise does not improve the baseline residual. Thus, only the Claude hypotheses improve both reported criteria substantially.

\begin{table}[H]
\centering
\caption{Single-run PINN results for the soliton with a Gaussian perturbation (problem without an analytical solution). The shared identity-baseline row contains the best value of each metric across the analyst runs; these values may therefore come from different runs. Improvements in \(\mathrm{Rel}_i\) are computed relative to this shared baseline.}
\label{tab:pinn_gauss_results}
\resizebox{\linewidth}{!}{
\begin{tabular}{@{\extracolsep{\fill}}lcccc@{}}
\toprule
Model & Result & \(\mathrm{Rel}_i\) & \(\mathrm{Res}_{\mathrm{bvp}}\) & Change in \(\mathrm{Rel}_i\) relative to baseline \\
\midrule
Human-designed PINN from prior work~\cite{zavertyaev2024opticalsolitons} & Reported result & \(9.87\cdot 10^{-4}\) & \(9.11\cdot 10^{-5}\) & \(130\times\) better \\
\midrule
\multicolumn{2}{l}{Identity baseline (best across runs)} & \(1.28\cdot 10^{-1}\) & \(1.04\cdot 10^{-3}\) & --- \\
\midrule
\texttt{sonar-pro} & Best \(\mathrm{Rel}_i\) & \(6.35\cdot 10^{-3}\) & \(2.37\cdot 10^{-2}\) & \(20\times\) better \\
\midrule
\texttt{anthropic-claude-opus-4.6} & Best \(\mathrm{Rel}_i\) & \(6.61\cdot 10^{-4}\) & \(3.95\cdot 10^{-5}\) & \(194\times\) better \\
\texttt{anthropic-claude-opus-4.6} & Best \(\mathrm{Res}_{\mathrm{bvp}}\) & \(1.40\cdot 10^{-3}\) & \(2.08\cdot 10^{-5}\) & \(91\times\) better \\
\midrule
\texttt{qwen-qwen3-30b-a3b-instruct-2507} & Best \(\mathrm{Rel}_i\) & \(9.43\cdot 10^{-2}\) & \(1.38\cdot 10^{-3}\) & \(1.36\times\) better \\
\bottomrule
\end{tabular}}
\end{table}

Ten additional trainings from random initializations were performed for the identity baseline and for both Claude configurations. The configuration selected by \(\mathrm{Res}_{\mathrm{bvp}}\) gives \(\mathrm{Rel}_i=(1.42\pm0.59)\cdot10^{-3}\), a \(116\times\) reduction from the repeated baseline mean, while the configuration selected by \(\mathrm{Rel}_i\) gives \((1.60\pm0.81)\cdot10^{-3}\), a \(103\times\) reduction. The repeated-run export does not contain \(\mathrm{Res}_{\mathrm{bvp}}\), so these series confirm robustness of the invariant error but not of the joint criterion. Details are provided in Appendix~\ref{app:pinn_extra}.

\subsection{PINN results: Kuramoto--Sivashinsky}
\label{sec:results_ks_gaussian}

In the traveling-wave problem with a Gaussian perturbation, the Claude Opus~4.6 agent generated the \(F_I,F_O\) transformations while minimizing the boundary-value-problem residual \(\mathrm{Res}_{\mathrm{bvp}}\). The mass drift \(\Delta_m^{(u)}\) was recorded alongside it but did not drive the search. The best values are summarized in Table~\ref{tab:ks_gauss_results}, and the two metrics are computed by different hypotheses.

On the primary metric (\(\mathrm{Res}_{\mathrm{bvp}}\)), the best architecture is \texttt{MultiplicativeIC\_DualAdditive\_DualZoom}, which reduces \(\mathrm{Res}_{\mathrm{bvp}}\) from the baseline value of \(1.41\cdot10^{-3}\) to \(1.511\cdot10^{-4}\), a factor of 9.3.
On the physical diagnostic, the best architecture is \texttt{BoundedMultAdd\_SechZoom\_DualPhase}, which reduces \(\Delta_m^{(u)}\) from \(9.31\cdot10^{-2}\) to \(6.56\cdot10^{-3}\), an approximately \(14\)-fold reduction, while also improving \(\mathrm{Res}_{\mathrm{bvp}}\) by a factor of 8.3.

\begin{table}[H]
\centering
\caption{Best single-run PINN results for the Kuramoto--Sivashinsky traveling wave with a Gaussian perturbation (Claude Opus~4.6, one seed per hypothesis). \(\mathrm{Res}_{\mathrm{bvp}}\) is the metric the analyst optimized; \(\Delta_m^{(u)}\) is an independent physical diagnostic. The two columns report the best value of each metric over the run; they are attained by different hypotheses, which are named in the last column.}
\label{tab:ks_gauss_results}
\resizebox{\linewidth}{!}{
\begin{tabular}{@{\extracolsep{\fill}}lccl@{}}
\toprule
Configuration & \(\mathrm{Res}_{\mathrm{bvp}}\) & \(\Delta_m^{(u)}\) & Hypothesis \\
\midrule
Baseline (\(F_I,F_O\) identity) & \(1.41\cdot 10^{-3}\) & \(9.31\cdot 10^{-2}\) & --- \\
Best \(\mathrm{Res}_{\mathrm{bvp}}\) (\(9\times\) better) & \(1.51\cdot 10^{-4}\) & \(3.47\cdot 10^{-2}\) & \texttt{MultiplicativeIC\_DualAdditive\_DualZoom} \\
Best \(\Delta_m^{(u)}\) (\(14\times\) better) & \(1.70\cdot 10^{-4}\) & \(6.56\cdot 10^{-3}\) & \texttt{BoundedMultAdd\_SechZoom\_DualPhase} \\
\bottomrule
\end{tabular}}
\end{table}

\begin{figure}
\centering
\includegraphics[width=\linewidth]{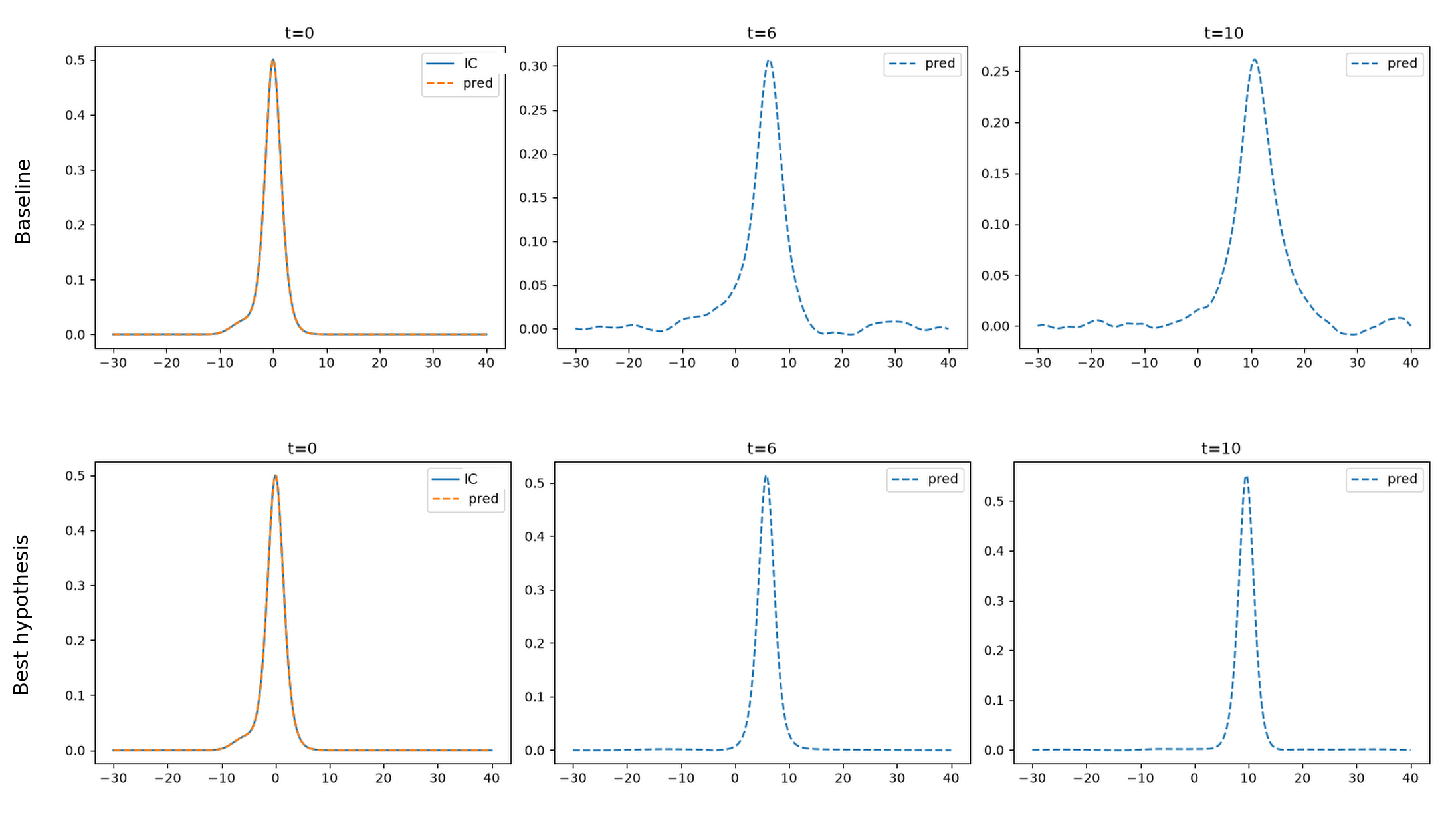}
\caption{Predicted solution of the Kuramoto--Sivashinsky problem at \(t=0,6,10\) for the baseline (top) and for the hypothesis with the lowest mass drift, \texttt{BoundedMultAdd\_SechZoom\_DualPhase} (bottom). No numerical reference is available for this problem, so only the PINN prediction is plotted; the initial condition is shown in the left panels for comparison. The horizontal axis is the coordinate \(x\), the vertical axis is \(u\). The baseline pulse loses roughly half of its amplitude by \(t=10\) and develops spurious low-amplitude oscillations across the domain, whereas the agent hypothesis keeps a localized traveling pulse whose amplitude stays close to the initial value of \(1/2\).}
\label{fig:ks_gauss_pred}
\end{figure}

Both winning hypotheses share a multiplicative--additive decomposition
\(\hat{u}=u(x,0)\,e^{-\alpha t}\bigl(1+t\,\gamma_m N_1\bigr)+t\,W(x)\bigl(\gamma_1 N_2+\gamma_2 N_3\bigr)\),
in which \(N_{1,2,3}\) are the three MLP outputs, the initial profile is embedded exactly at \(t=0\), the window \(W(x)=s_L s_R\) built from the two normalized distances to the boundaries vanishes at \(x=-30\) and \(x=40\), and the input block supplies features in the co-moving coordinate \(\xi=x-t\) --- a rescaled \(\tanh(\xi/\lambda)\), the profile \(\mathrm{sech}^2(\xi/2)\) of the unperturbed pulse, and a phase marker \(\sin(\pi t/6)\) of the collision. The hypothesis that wins on mass drift differs in one detail: the multiplicative channel is bounded, \(N_1\mapsto\tanh(N_1)\), so the correction cannot rescale the embedded initial profile without limit. This bound costs little in terms of the residual but keeps the integral quantity \(M=\int u\,dx\) an order of magnitude closer to its initial value, which illustrates that the two criteria are not interchangeable.

\section{Ablation studies}
\label{sec:ablation}

\subsection{Simple and detailed prompts}

This ablation compares the simple and detailed SNN prompt configurations defined in Section~\ref{sec:setup_prompts}. The detailed configuration combines the method-class taxonomy, structured Chain-of-Thought, and coverage map described by the general framework in Section~\ref{sec:agent}. The full Analyst and Coder prompts are provided in Supplementary Section~\ref{app:prompts}.

The MNIST configurations are compared in Table~\ref{tab:prompt_model_configs}. For Claude Opus~4.6, the detailed prompt increased the mean maximum accuracy over ten matched complete runs from \(91.13\pm0.44\,\%\) to \(92.15\pm0.62\,\%\), and the best result from 91.65\,\% to 92.83\,\%. The difference between the prompt variants is confirmed by a two-sided exact Wilcoxon signed-rank test (\(W=5\), \(p=0.0195\)), with the detailed prompt yielding higher accuracy in eight of ten pairs; the mean paired difference is \(1.02\pm0.90\)~percentage points (Table~\ref{tab:opus_prompt_paired_runs}), meaning that the effect is comparable in magnitude to the between-run variation.

Ten complete runs were performed with each prompt variant for Perplexity Sonar and Qwen3-30B-A3B-Instruct-2507. With the simple prompt, none of the 250 decoders proposed per run outperformed the baseline, so the run maximum for both models exactly matched the baseline of 79.88\,\% in all ten cases. The detailed prompt changed the results of both models, but to different extents. The 90\,\% threshold was fixed before the run series as a reference point for MNIST; for the \texttt{digits} dataset, where the baseline is already 90.83\,\%, an analogous reference point was set at 93\,\% Section~\ref{sec:setup_eval} and is not used in the tables in this section. For Perplexity Sonar, the individual-run maxima ranged from 90.93\,\% to 92.70\,\% (Table~\ref{tab:sonar_prompt_paired_runs}), with a mean of \(91.76\pm0.56\,\%\), so the 90\,\% threshold was exceeded in all ten runs. For Qwen3-30B-A3B-Instruct-2507, the mean maximum was \(85.88\pm2.42\,\%\), and the best run reached 89.16\,\% without exceeding the specified 90\,\% threshold. The difference between the prompt variants for both models is confirmed by a two-sided exact paired Wilcoxon signed-rank test.

A run-by-run comparison between the simple and detailed prompts for Qwen3-30B-A3B-Instruct-2507 is provided in Table~\ref{tab:qwen_prompt_paired_runs}. With the simple prompt, accuracy remained at the baseline level of $79.88\,\%$ in all ten runs. The detailed prompt provided gains ranging from $2.40$ to $9.28$ percentage points in individual runs. The mean maximum accuracy was $85.88 \pm 2.42\,\%$, and the best result was $89.16\,\%$. The high variability ($\mathrm{std}=2.42$ percentage points at $n=10$) indicates instability of the search when this model is used.

\subsection{Coverage-map ablation}

The ablation experiment excluding the coverage map was performed for all three analyst models (Tables~\ref{tab:qwen_ablation_prompt_paired_runs},~\ref{tab:sonar_ablation_prompt_paired_runs}, and~\ref{tab:opus_ablation_prompt_paired_runs}). For Claude Opus~4.6, removing the map reduces the mean maximum to \(91.02\pm0.25\,\%\), which is \(1.14\) percentage points below the detailed prompt (\(92.15\,\%\)) and nearly matches the level of the simple prompt (\(91.13\,\%\)). All ten runs nevertheless retain accuracy above 90\,\%, indicating that the coverage map has little effect on the result for the strong model. For Perplexity Sonar, removing the map reduces the mean maximum to \(82.34\pm0.70\,\%\) ($\Delta=9.42$ percentage points), and no run reaches the threshold. For Qwen3-30B-A3B-Instruct-2507, the mean maximum falls to \(80.18\pm0.45\,\%\) ($\Delta=5.70$ percentage points), effectively returning to the baseline level of 79.88\,\%: the coverage map proves to be the key component that enabled this model to improve over the baseline argmax decoder (the class label is assigned by the neuron with the highest spike rate).

\section{Discussion}
\label{sec:discussion}

The experiments show that prompt detail can partially compensate for the limitations of the analyst model, although the degree of compensation depends on the LLM used. Claude Opus~4.6 achieves accuracy above 90\,\% with both prompt variants. With the simple prompt, Perplexity Sonar remains at the baseline level in all ten complete runs, but with the detailed prompt it consistently exceeds 90\,\% and reaches the level of Claude Opus~4.6. For Qwen3-30B-A3B-Instruct-2507, the detailed prompt also produces a substantial gain: the best result increases from 79.88\,\% to 89.16\,\%. At the same time, the higher variability of the maxima (\(85.88\pm2.42\,\%\)) and the absence of any run with a maximum accuracy above 90\,\% show that the instruction structure does not fully compensate for the limitations of this model. When search cost is taken into account (Table~\ref{tab:cost_summary_mnist}), the advantage of Claude Opus~4.6 also persists in economic terms: despite the highest cost per run (\$98 versus \$16 for Sonar), it reaches its best decoder after only \(k_{\min}=2\) to \(k_{\max}=43\) evaluated candidates (Table~\ref{tab:tpot_vs_agent}), which makes the cost of \emph{reaching the best result} comparable to that of Sonar (\$0.37 versus \$0.21), whereas Qwen3-30B-A3B-Instruct-2507, deployed on a rented GPU, costs less than both proprietary models (\$5.90 per run at the managed rate for the same model class, with the Analyst alone contributing \$2.00) but still fails to reach the target threshold over the full run.

In every task, the agent finds a technique tied to the physics of the problem: class-conditional spike-rate statistics for the CCN, the soliton structure of the solution for the Schrödinger equation, and a phase-separated representation of the pulse--perturbation interaction for the Kuramoto--Sivashinsky equation. The second PINN problem (a soliton with a Gaussian perturbation) is particularly informative: its analytical solution is unknown, so the gain cannot be reduced to substituting a ready-made formula, and the best configuration (Claude Opus~4.6) improves the conservation of the integrals of motion by more than an order of magnitude. In the KS problem, the improvement appears in a different physical diagnostic, namely mass drift. All five experiments are nevertheless handled by the same framework without changes to its internal structure.

\section{Limitations}
\label{sec:limitations}

Despite the results reported above, the present study has several
limitations that should be taken into account when interpreting the
findings and when comparing our system with alternative approaches.

\begin{itemize}
    \item \textbf{Restricted search space.} The agent modifies a fixed
    catalogue of editable blocks (encoders, neuron models, connectivity
    schemes for SNNs; loss weights, activation functions and sampling
    strategies for PINNs). Structural transformations that fall outside
    this catalogue -- for example, adding auxiliary networks or
    replacing the entire training paradigm -- cannot be discovered by
    the current agent and would require an extended action space.

    \item \textbf{Flat experiment log without a hypothesis graph.} The
    Recorder stores runs as a chronological JSON log; it does not
    maintain a directed acyclic graph of hypotheses and their
    dependencies. As a consequence, the Analyst re-derives relations
    between experiments from the log at every step, which limits the
    depth of multi-step reasoning that can be sustained over long
    sessions.

    \item \textbf{Bounded number of iterations.} All reported runs are
    capped at $250$ iterations for SNN tasks and $50$ iterations for
    PINN tasks. These bounds were chosen to keep the wall-clock cost
    reproducible within the budgets reported in Table~\ref{tab:cost_summary_mnist},
    but they also mean that the accuracy and relative-error values in
    Sections~\ref{sec:results_snn}, \ref{sec:results_pinn_one}, \ref{sec:results_pinn_gauss} and \ref{sec:results_ks_gaussian} are lower
    bounds on what the agent can eventually reach with a larger budget.

    \item \textbf{Single-seed evaluation in part of the regimes.} For
    the most expensive configurations (Claude Opus~4.6 on the
    generalised nonlinear Schr\"odinger equation and on the
    Kuramoto--Sivashinsky benchmark) we report a single seed per
    configuration; the ten-run averages for best models are given only for the soliton with a Gaussian perturbation case. A full multi-seed sweep for every
    LLM/task pair is left for future work.

    \item \textbf{Dependence on proprietary LLM APIs.} Two of the three
    evaluated models (Claude Opus~4.6 and Perplexity Sonar) are
    accessed through commercial APIs; their availability, pricing and
    behaviour may change between the time of the experiments and the
    time of reproduction. The open-weights Qwen3-30B-A3B run is
    provided as a partial mitigation, but its cost/quality trade-off is
    not directly comparable to the proprietary systems.

    \item \textbf{No head-to-head comparison with LLM-based NAS
    baselines.} The reported baselines are a random-choice controller
    and the state-of-the-art results reported in the cited references.
    A direct comparison with recent LLM-driven neural-architecture-search
    systems on the same tasks and under the same iteration and cost
    budgets is desirable but was outside the scope of the present
    paper; we consider it the natural next step of this line of
    research.
\end{itemize}

\section{Conclusion}
\label{sec:conclusion}

To date, automatic tuning of neural networks has been carried out mainly within the framework of an allocated single  metric of the given competition: agent-based systems optimize the entire ML script, NAS approaches search over architectures according to accuracy, and PINN frameworks train networks to minimize the relative error. For tasks whose quality is defined by a multidimensional signal---by the conservation of a set of physical invariants or by additional structural requirements---frameworks of a different class are required. The results reported in this paper demonstrate the robustness and generalization ability of the proposed multi-agent framework, which consists of an analyst, a programmer, and a recorder, together with a sandboxed execution environment and an experiment log serving as external memory, on tasks of different nature without changes to its internal structure.

The same unchanged framework was applied to five tasks with different types of information representation: accuracy increased from 79.88\,\% to 92.83\,\% on MNIST and from 90.83\,\% to 96.67\,\% on \texttt{digits}; in the single-soliton PINN problem, the relative amplitude error decreased by a factor of approximately 2491; in the more complex problem with a Gaussian perturbation and no known analytical solution, the conservation error of the integrals of motion decreased by a factor of approximately 325 in the selected run and by factors of 116 and 103, respectively, based on the mean errors over ten additional trainings of the residual-selected and invariant-selected configurations (Table~\ref{tab:pinn_repeats}); and in the Kuramoto--Sivashinsky problem, the boundary-value-problem residual decreased by a factor of 9.3 and mass drift by a factor of 14 (Tables~\ref{tab:summary} and~\ref{tab:ks_gauss_results}). A separate result concerns the interaction between the analyst model and prompt structure. On MNIST, the detailed prompt enabled Perplexity Sonar to exceed 90\,\% accuracy in all ten complete runs and to reach a maximum of 92.70\,\%, close to the result of Claude Opus~4.6. For Qwen3-30B-A3B-Instruct-2507, maximum accuracy increased from the baseline level of 79.88\,\% to 89.16\,\%, but the target threshold was not reached. Prompt design can therefore substantially improve the performance of a less capable LLM, but only partially compensates for differences between models.

Future work includes increasing the complexity of the PINN problem (the interaction of two solitons and other regimes without analytical solutions), jointly tuning several blocks (the decoder and plasticity rules for the SNN), and constructing models that account for the constraints of neuromorphic systems.

\section*{Declaration of competing interest}
The authors declare that they have no known competing financial interests or personal relationships that could have appeared to influence the work reported in this paper.

\section*{Funding}
This work was supported by ongoing institutional funding. No additional grants to carry out or direct this particular research were obtained.

\section*{Acknowledgments}

The work was carried out according to the state assignment of the National Research Centre "Kurchatov Institute" using the computing resources of the federal center for collective use "Complex for modeling and data processing for megascience class installations" of the National Research Center "Kurchatov Institute", \url{http://ckp.nrcki.ru/}.

\section*{Data availability}

The results of the SNN and PINN framework will be made publicly available in the~\url{https://github.com/sag111/llm-auto-tuning-results}.

Additionally, the full texts of the hypotheses proposed by the LLM Analyst and the code written by the LLM Coder for these hypotheses are presented there.
The full texts of the prompts are available upon request to the authors of the article.

\section*{Declaration of generative AI and AI-assisted technologies in the writing process}

During the preparation of this work, the authors used Perplexity AI to assist with language editing and improving the clarity of the manuscript. After using this tool, the authors reviewed and edited the content as needed and take full responsibility for the content of the manuscript.

The large language models evaluated in this study were also used as research tools within the experimental methodology described in Sections~\ref{sec:agent} and~\ref{sec:protocol}. This methodological use is distinct from their use in manuscript preparation.

\printcredits

\appendix

\section{Baseline PINN model}
\label{app:pinn}

The model problem is a boundary-value problem for the generalized nonlinear Schrödinger equation (GNLSE) with cubic--quintic nonlinearity:
\begin{equation}
\begin{cases}
i\, q_t + q_{xx} + |q|^{2} q\,\bigl(1 - \alpha|q|^{2} + \beta|q|^{4}\bigr) = 0,\\[4pt]
q(x,\,0) = \tilde{q}(x,\,0),\\[2pt]
q(x_{\min},\,t) = 0, \quad q(x_{\max},\,t) = 0,\\[2pt]
(x,t)\in[x_{\min},x_{\max}]\times[t_{\min},t_{\max}].
\end{cases}
\label{eq:gnlse}
\end{equation}
In all configurations, \(\beta = 0\); the quintic term is retained in the equation for generality. Two configurations are considered. In the single-soliton problem, \(\alpha = 0.1\) and the domain is \((x,t)\in[-25,25]\times[0,3]\); the initial profile is specified by the analytical soliton solution:
\begin{equation}
\tilde{q}(x,0) =
\sqrt{
\frac{\mu z}
{\bigl(\tfrac{1}{2}z + 1\bigr)^{2} - \tfrac{\alpha\mu}{3}z^{2}}
}
\cdot e^{i(kx + \theta_0)},
\qquad
z=e^{\sqrt{\mu}(x-x_0)},
\label{eq:q0}
\end{equation}
where \(k=1\), \(w=0.88\), \(x_0=0\), \(\theta_0=0\) and \(\mu=4(k^{2}-w)\). The complex field is decomposed into two real components: \(q(x,t) = u(x,t) + i\,v(x,t)\). The analytical soliton is used as the reference when calculating the error.

In the second configuration (a soliton with a Gaussian perturbation), \(\alpha = 0.3\), the domain is extended to \((x,t)\in[-50,50]\times[0,25]\), and the initial condition is the sum of the soliton profile~\eqref{eq:q0} (with \(x_0=-30\)) and a real-valued Gaussian packet:
\begin{equation}
q(x,0) = q_{2}(x,0) + 0.2\,e^{-x^{2}/50},
\end{equation}
where \(q_{2}\) coincides with~\eqref{eq:q0} for the specified parameters, equivalently \(A_g\exp[-(x-B_g)^{2}/(2C_g^{2})]\) with \(A_g=0.2\), \(B_g=0\), and \(C_g=5\) (the printed \(C_g^{2}\) in~\cite{zavertyaev2024opticalsolitons} is a typographical error, must be \(2C_g^{2}\)). The analytical solution is unknown. A PINN field from~\cite{zavertyaev2024opticalsolitons} is retained as an auxiliary amplitude reference. On \([-50,50]\times[0,25]\) that field has \(\mathrm{Res}_{\mathrm{bvp}}=9.11\cdot10^{-5}\) and \( \mathrm{Rel}_i=9.87\cdot10^{-4}\). The baseline network in this configuration is deeper, with 4 hidden layers of 64 neurons each, and collocation-point sampling is adaptive (version v3 from~\cite{moloshnikov2025solitonmethods}, regenerated every 1000 iterations with coefficient \(\lambda_2=0.01\) in $P_rl$ in Eq. 6 of~\cite{moloshnikov2025solitonmethods}).

The primary metrics of the search are \(\mathrm{R es}_{\mathrm{bvp}}\)~\eqref{eq:gnlse_bvp_residual} and the relative conservation error of the integrals of motion \(\mathrm{Rel}_i = \mathrm{Rel}_{i_1} + \mathrm{Rel}_{i_2}\); \(\mathrm{Rel}_h\) has lowest priority.

The baseline is a fully connected network \(\mathrm{MLP}\colon \mathbb{R}^2 \to \mathbb{R}^2\), \((x,t)\mapsto(u,v)\), with two hidden layers of 32 neurons each and \(\sin\) activation. In the modified architecture, two tunable blocks, \(F_I(x,t)\) and \(F_O(u,v,x,t)\), are attached to the MLP; both blocks are identity mappings in the baseline. The complete set of hyperparameters for both configurations is summarized in Table~\ref{tab:hyperparams}. The loss functional consists of three terms with equal weights:
\begin{equation}
\mathcal{L} = \mathcal{L}_{\mathrm{pde}} + \mathcal{L}_{\mathrm{ic}} + \mathcal{L}_{\mathrm{bc}},
\label{eq:loss}
\end{equation}
where \(\mathcal{L}_{\mathrm{pde}}\) is the mean squared residual of Eq.~\eqref{eq:gnlse} at the collocation points, \(\mathcal{L}_{\mathrm{ic}}\) is the initial-condition error, and \(\mathcal{L}_{\mathrm{bc}}\) is the boundary-condition error. The derivatives \(q_t,q_{xx}\) are obtained by TensorFlow automatic differentiation. For the single-soliton problem the metric is:
\begin{equation}
\mathrm{Rel}_h = \frac{\|h_{\mathrm{true}} - h_{\mathrm{pred}}\|_2}{\|h_{\mathrm{true}}\|_2}, \qquad h = \sqrt{u^2 + v^2}.
\label{eq:relh}
\end{equation}

\begin{table}[htbp]
\centering
\caption{Hyperparameters of the baseline PINN for the two GNLSE configurations. The Kuramoto--Sivashinsky configuration is given separately in Table~\ref{tab:hyperparams_ks}.}
\label{tab:hyperparams}
\begin{tabular}{lcc}
\toprule
Parameter & Single soliton & Soliton + Gaussian \\
\midrule
Number of hidden layers & 2 & 4 \\
Neurons per layer & 32 & 64 \\
Activation (hidden layers) & \(\sin\) & \(\sin\) \\
Activation (output layer) & linear & linear \\
Optimizer & Adam & Adam \\
Learning rate & \(10^{-3}\) & \(10^{-3}\) \\
Number of training iterations & 10\,000 & 30\,000 \\
Collocation points (\(N_f\)) & 1\,000 & 30\,000 \\
Initial-condition points (\(N_0\)) & 100 & 3\,000 \\
Boundary-condition points (\(N_b\)) & 50 & 300 \\
Collocation-point sampling & fixed random & adaptive (v3) \\
Weights \(w_\mathrm{pde},\,w_\mathrm{ic},\,w_\mathrm{bc}\) & \(1, 1, 1\) & \(1, 1, 1\) \\
\bottomrule
\end{tabular}
\end{table}

\subsection{Kuramoto--Sivashinsky baseline hyperparameters}
\label{app:ks_hyperparams}

The Kuramoto--Sivashinsky problem uses its own baseline configuration, summarized in Table~\ref{tab:hyperparams_ks}. It differs from the GNLSE configurations in three respects. First, the network output is a single real scalar. Second, the fourth derivative in the residual makes the raw loss stiff, so the residual is divided by \(R_0=10\) before squaring, the learning rate is lowered to \(5\cdot10^{-5}\), and the gradient norm is clipped at 1.
Without these three measures a substantial fraction of initializations diverges to non-finite loss within the first few hundred iterations. Third, the boundary term penalizes both \(u\) and \(u_{xx}\) at the two boundaries, so the loss reads
\[
\mathcal{L}
=\bigl\langle(\mathcal{R}_{\mathrm{KS}}/R_0)^2\bigr\rangle
+\bigl\langle(u_{\mathrm{pred}}(x,0)-u(x,0))^2\bigr\rangle
+\bigl\langle u^2+u_{xx}^2\bigr\rangle_{\partial}.
\]

\begin{table}[htbp]
\centering
\caption{Hyperparameters of the baseline PINN for the Kuramoto--Sivashinsky traveling-wave problem with a Gaussian perturbation.}
\label{tab:hyperparams_ks}
\begin{tabular}{lc}
\toprule
Parameter & Value \\
\midrule
Number of hidden layers & 4 \\
Neurons per layer & 32 \\
Network output & \(u\) (one real scalar) \\
Activation (hidden layers) & \(\sin\) \\
Activation (output layer) & linear \\
Optimizer & Adam, \(\mathrm{clipnorm}=1\) \\
Learning rate & \(5\cdot10^{-5}\) \\
Residual scale \(R_0\) & 10 \\
Number of training iterations & 50\,000 \\
Collocation points (\(N_f\)) & 20\,000 \\
Initial-condition points (\(N_0\)) & 1\,000 \\
Boundary-condition points (\(N_b\)) & 200 \\
Collocation-point sampling & adaptive (v3), every 1\,000 iterations \\
Adaptive-sampling parameters & \(\lambda_2=0.01\), oversampling \(\times5\) \\
Evaluation grid (\(n_x\times n_t\)) & \(1000\times10\) \\
Weights \(w_\mathrm{pde},\,w_\mathrm{ic},\,w_\mathrm{bc}\) & \(1, 1, 1\) \\
\bottomrule
\end{tabular}
\end{table}

\subsection{Kuramoto--Sivashinsky field shift and PINN residual}
\label{app:ks_shift}

For completeness, we derive the zero-background form of the generalized Kuramoto--Sivashinsky equation used by the PINN. We use the normalization
\begin{equation}
v_t+v_{xx}+\sigma v_{xxx}+v_{xxxx}
+\gamma v v_x+\chi(v^2)_{xx}=0,
\label{eq:ks_appendix_source}
\end{equation}
where \(\gamma=2\beta\), so that the nonlinear convection term integrates to \(\beta v^2\) under a traveling-wave reduction. For \(\beta=3\sigma\), substitution of \(\sigma=-1\), \(\gamma=2\beta=6\sigma=-6\), and \(\chi=3\) gives
\begin{equation}
v_t+v_{xx}-v_{xxx}+v_{xxxx}
-6v v_x+3(v^2)_{xx}=0.
\label{eq:ks_appendix_parameters}
\end{equation}

The exact wave from~\cite{kudryashov2026traveling} is expressed on the background \(-1/3\). To make the learned field decay to zero away from the pulse, we set
\begin{equation}
v=u-\frac{1}{3},
\qquad
u=v+\frac{1}{3}.
\label{eq:ks_field_shift}
\end{equation}
Because the shift is constant, all differentiated linear terms are unchanged:
\[
v_t=u_t,\quad v_x=u_x,\quad v_{xx}=u_{xx},\quad
v_{xxx}=u_{xxx},\quad v_{xxxx}=u_{xxxx}.
\]
The nonlinear convection term transforms as
\begin{equation}
-6v v_x
=-6\left(u-\frac{1}{3}\right)u_x
=-6u u_x+2u_x,
\label{eq:ks_convection_shift}
\end{equation}
whereas the nonlinear diffusion term becomes
\begin{align}
v^2
&=\left(u-\frac{1}{3}\right)^2
=u^2-\frac{2}{3}u+\frac{1}{9},\\
3(v^2)_{xx}
&=3(u^2)_{xx}-2u_{xx}.
\label{eq:ks_diffusion_shift}
\end{align}
Inserting these expressions into Eq.~\eqref{eq:ks_appendix_parameters} yields
\[
u_t+u_{xx}-u_{xxx}+u_{xxxx}
-6u u_x+2u_x+3(u^2)_{xx}-2u_{xx}=0.
\]
Combining the two \(u_{xx}\) terms gives the residual used in the implementation:
\begin{equation}
\mathcal{R}_{\mathrm{KS}}
=u_t-u_{xx}-u_{xxx}+u_{xxxx}
-6u u_x+2u_x+3(u^2)_{xx}.
\label{eq:ks_pinn_residual}
\end{equation}
Thus, the negative coefficient of \(u_{xx}\) and the additional linear-advection term \(2u_x\) are both consequences of removing the constant background; they are not independent changes to the source equation.

For the unperturbed case, the source-field solution and its zero-background counterpart are
\[
v_{\mathrm{ex}}(x,t)
=-\frac{1}{3}
+\frac{1}{2\cosh^2((x-t-x_0)/2)},
\qquad
u_{\mathrm{ex}}(x,t)
=\frac{1}{2\cosh^2((x-t-x_0)/2)}.
\]

\section{Baseline SNN model}
\label{app:snn}

The Correlation Classwise Network (CCN)~\cite{sboev2024direct,kunitsyn2025probabilistic} was selected as the baseline model. Its architecture, input encoding, supervisory-signal delivery, and plasticity rule are adopted without modification; the agent's modifications are restricted to the output-spike-activity decoder. Input expansion using Gaussian receptive fields (GRF) and the bagging ensemble~\cite{rybka2023image} are not used because preliminary experiments on MNIST showed no appreciable gain.

The approach is applied to two datasets: MNIST (\(28\times28\), \(N=784\) spike generators) and the scikit-learn \texttt{digits}~\cite{scikit-learn} (\(8\times8\), \(N=64\)). Both have ten classes, pixel values are normalized by the \(L_1\) norm of the input vector, and the number of output LIF neurons is \(K=10\). The network has two layers: the input layer contains \(N\) spike generators, and the output layer contains \(K\) Leaky-Integrate-and-Fire (LIF) neurons; the connections are fully connected and trainable, with no lateral connections.

The input vector \(V=(v_1,\dots,v_N)\) is converted into an array of correlated spike sequences \(\{S_i\}\): a reference sequence \(S_0\) of length \(N_t\) is constructed with constant spacing \(N_{\mathrm{ref}}\); component \(v_i\) is interpreted as the target Pearson correlation coefficient between \(S_i\) and \(S_0\), and \(S_i\) is obtained by randomly removing some spikes from \(S_0\). Supervision is implemented using two currents applied to the output LIF neurons: for a sample of class \(k\), a positive current \(I_{\mathrm{exc}}=10^{7}\) pA, synchronized with the \(S_0\) spikes with a shift \(\tau_s\), is applied to the corresponding neuron; the other neurons receive an inhibitory current \(I_{\mathrm{inh}}=-10^{3}\) pA. The weights are adjusted using a local multiplicative STDP rule with a bounding symmetric nearest-neighbor scheme:
\[
\Delta w(\Delta t)=
\begin{cases}
A^{+}\cdot w^{\mu_+} \cdot e^{-\Delta t/\tau_{+}}, & \Delta t>0,\\
A^{-}\cdot w^{\mu_-} \cdot e^{\Delta t/\tau_{-}}, & \Delta t<0,
\end{cases}
\]
where \(\Delta t = t_{\text{post}} - t_{\text{pre}}\), \(w\) is the current weight, \(A^{\pm}\) are the potentiation and depression amplitudes, \(\tau_{\pm}\) are the window time constants, and \(\mu_{\pm}\) are the exponents of the weight dependence. The values of all network and STDP-rule parameters are provided in Table~\ref{tab:hyperparams_snn}.

\begin{table}[htbp]
\centering
\caption{Hyperparameters of the baseline SNN and its training, common to all configurations.}
\label{tab:hyperparams_snn}
\begin{tabular}{ll}
\toprule
Parameter & Value \\
\midrule
\(V_{rest}\) (resting potential) & \(-70\) mV \\
\(V_{reset}\) (post-spike potential) & \(-70\) mV \\
\(V_{th}\) (threshold potential) & \(-56\) mV \\
\(\tau_m\) (membrane time constant) & \(10^{3}\) ms\\
\(\tau_s\) (teacher-spike shift) & \(0.1\) ms\\
\(N_{ref}\) (spacing between spikes in \(S_0\)) & 3 \\
\(N_{t}\) (number of time steps in \(S_0\)) & 3000 \\
\(T_{sim}\) (simulation time, \(dt=0.1\) ms) & 300 ms \\
\(A^{+}\) (potentiation amplitude) & \(1.2 \cdot 10^{-3}\) \\
\(A^{-}\) (depression amplitude) & \(10^{-3}\) \\
\(\tau_{+}\) (potentiation window) & \(200\) ms\\
\(\tau_{-}\) (depression window) & \(160\) ms\\
\(\mu_+\) (weight dependence, potentiation) & \(0.96\) \\
\(\mu_-\) (weight dependence, depression) & \(0.26\) \\
\bottomrule
\end{tabular}
\end{table}

\subsection{SNN experimental protocol}
\label{sec:snn_protocol}

\paragraph{MNIST data split.} The standard MNIST test set, containing 10\,000 samples, was not used during the agent search. The original training set of 60\,000 samples was stratified into training and validation subsets at a \textbf{80:20} ratio. The parameters of each proposed decoder were estimated on the training subset, whereas the classification accuracy passed to the agent-search log was calculated only on the validation subset.

After the specified search budget was exhausted, the decoder with the highest validation accuracy was selected. The selected decoder was retrained on the complete original MNIST training set and evaluated once on the standard test set. Test samples, their spike representations, and their labels were not provided to the agents and were not used to select the decoder structure or tune its hyperparameters. The same stratified split of the training set was used in all compared configurations.

\paragraph{\texttt{digits} data split.} Before the agent search began, the \texttt{digits} dataset, containing 1797 samples, was stratified into a development set of 1437 samples and a held-out test set of 360 samples. The test set was not used to calculate metrics passed to the agents, select the decoder structure, or tune its hyperparameters.

Stratified five-fold cross-validation was performed on the development set. The five-fold split was fixed once before the experiments and remained unchanged for all decoders, runs, analyst models, prompt variants, and baseline methods. At each iteration, the proposed decoder was trained five times on four folds and evaluated on the remaining fold. The mean classification accuracy was recorded in the experiment log:
\begin{equation}
    \operatorname{CVAcc}(d)
    =
    \frac{1}{5}
    \sum_{k=1}^{5}
    \operatorname{Acc}
    \left(
        d^{(-k)}, \mathcal{D}^{(k)}
    \right),
    \label{eq:digits_cv_accuracy}
\end{equation}
where \(d^{(-k)}\) denotes the decoder trained on four folds, and \(\mathcal{D}^{(k)}\) is the fifth fold excluded from training. The analyst used \(\operatorname{CVAcc}(d)\) when formulating the next hypothesis.

After the specified search budget was exhausted, the decoder with the highest \(\operatorname{CVAcc}(d)\) was selected. It was then retrained on all 1437 samples in the development set and evaluated once on the 360 samples in the held-out test set. The resulting test accuracy was not recorded in the agent-search log.

\section{Analysis of the best PINN hypotheses}
\label{app:pinn_analysis}

The single-soliton comparison includes Perplexity Sonar (\texttt{sonar-pro}), Claude Opus~4.6, and Qwen3-30B-A3B-Instruct-2507 (Table~\ref{tab:pinn_one_results}). The agent controls the input transformation \(F_I\) and output transformation \(F_O\), while the MLP, activation function, collocation points, loss functional, and training schedule remain fixed. The best overall result was obtained by the Claude Opus~4.6 configuration \texttt{PCDS-LCC-EXP-NLOG-MIXCROSS}, which gives \(\mathrm{Rel}_h=2.21\cdot10^{-5}\), a \(2491\)-fold reduction from its identity baseline of \(5.51\cdot10^{-2}\). It uses seven analytical features, including normalized and logarithmic envelopes and two amplitude--time cross-features, while \(F_O\) applies time-gated exponential amplitude and additive phase corrections with separate envelope weights.

The best \texttt{sonar-pro} configuration, \texttt{tw\_nonlinear\_pinn\_core}, gives \(\mathrm{Rel}_h=1.97\cdot10^{-4}\), a \(238\)-fold reduction from its identity baseline of \(4.69\cdot10^{-2}\). Its \(F_I\) supplies a normalized traveling-wave coordinate, normalized time, and a local nonlinearity-density feature; \(F_O\) uses the analytical traveling-wave amplitude and phase and interprets the MLP outputs as bounded amplitude and phase corrections. The best Qwen configuration, \texttt{LogPhaseHybridNorm}, gives \(\mathrm{Rel}_h=8.86\cdot10^{-3}\), a \(5.48\)-fold reduction from its identity baseline of \(4.85\cdot10^{-2}\). It normalizes the analytical soliton coordinate at the input and reconstructs the field from the analytical amplitude and carrier phase multiplied by learned complex corrections.

All three best configurations exploit the available analytical solution. Their performance should therefore be interpreted as evidence that the framework can discover a useful physics-informed representation, rather than as evidence of performance on problems without a known solution.

\section{Repeated PINN runs and statistical analysis}
\label{app:pinn_extra}
To assess whether the observed improvements are robust to random initialization, we repeated the training procedure for the GNLSE problem with a Gaussian perturbation. We compared the baseline model with two agent-designed configurations: one selected for its low equation residual and the other for its accurate preservation of the physical invariant. Each configuration was successfully trained ten times. Because the random initializations were not matched across configurations, the resulting samples were treated as independent rather than paired. Table~\ref{tab:pinn_repeats} summarizes the results.
\begin{table}[htbp]
\centering
\caption{Repeated runs for the GNLSE soliton problem with a Gaussian perturbation. The table reports the mean and sample standard deviation, median, range, and coefficient of variation (CV) of \(\mathrm{Rel}_i\).}
\label{tab:pinn_repeats}
\resizebox{\linewidth}{!}{
\begin{tabular}{@{\extracolsep{\fill}}lccccc@{}}
\toprule
Configuration & Mean \(\pm\) std & Median & Min--max & CV & \makecell{Reduction factor in \(\mathrm{Rel}_i\)\\relative to baseline mean} \\
\midrule
Baseline (\(n=10\)) & \((1.65\pm0.18)\cdot10^{-1}\) & \(1.66\cdot10^{-1}\) & \(1.29\cdot10^{-1}\)--\(1.94\cdot10^{-1}\) & \(11.1\,\%\) & \(1\times\) \\
Residual-selected configuration (\(n=10\)) & \((1.42\pm0.59)\cdot10^{-3}\) & \(1.50\cdot10^{-3}\) & \(5.40\cdot10^{-4}\)--\(2.41\cdot10^{-3}\) & \(41.8\,\%\) & \(116.1\times\) \\
Invariant-selected configuration (\(n=10\)) & \((1.60\pm0.81)\cdot10^{-3}\) & \(1.59\cdot10^{-3}\) & \(7.57\cdot10^{-4}\)--\(3.31\cdot10^{-3}\) & \(50.4\,\%\) & \(103.3\times\) \\
\bottomrule
\end{tabular}}
\end{table}

Both agent-designed configurations reduce the invariant error by more than two orders of magnitude on average relative to the baseline. The residual-selected configuration has a slightly lower mean error and coefficient of variation, although the two distributions overlap substantially. In the original search, the invariant-selected configuration achieved an error of \(6.61\cdot10^{-4}\), lower than in any of its ten subsequent retrainings. By contrast, the original error of the residual-selected configuration, \(1.40\cdot10^{-3}\), is close to its repeated-run mean. This contrast indicates a selection effect: the best result found during a search can be more favorable than the performance typically obtained when the same configuration is retrained.

\section{Additional classification results}
\label{app:snn_extra}

\paragraph{Method-class coverage map.} The aggregate coverage map of decoder method classes from the complete 250-iteration log for the \(\langle\)Perplexity Sonar, detailed prompt\(\rangle\) configuration on MNIST is presented in Table~\ref{tab:method_class_coverage_250}. Each valid decoder is assigned to one principal class according to the substance of its decision rule, while decoders that explicitly aggregate several predictions are grouped in the ``Ensembles'' row. The map shows that the analyst spent most of the budget on hybrid ensemble schemes and discriminant analysis, whereas the correlation, margin-based, and argmax classes remained uncovered in this run.

\begin{table}[htbp]
\centering
\caption{Coverage map of decoder method classes for one 250-iteration run
(Perplexity Sonar, detailed prompt, MNIST).}
\label{tab:method_class_coverage_250}
\begin{tabular}{@{}p{0.40\linewidth}cc>{\raggedright\arraybackslash}p{0.22\linewidth}@{}}
\toprule
Method class & Methods & Best acc, \% & Status \\
\midrule
Ensembles                       & 154 & 92.70 & explored \\
LDA / QDA                       &  32 & 87.90 & explored \\
Softmax / calibration           &  13 & 82.32 & explored \\
\(k\)-nearest neighbors         &  11 & 88.56 & explored \\
Naive Bayes                     &  11 & 83.73 & explored \\
Mahalanobis                     &   9 & 86.05 & limited data \\
Rank-based                      &   8 & 81.68 & limited data \\
Prototypes / centroids          &   5 & 80.43 & limited data \\
KDE / density                   &   4 & 85.75 & limited data \\
Threshold methods / rules       &   3 & 73.93 & limited data \\
Correlation / cosine            &   0 & ---   & not explored \\
Margin-based                    &   0 & ---   & not explored \\
\bottomrule
\end{tabular}
\end{table}

\paragraph{Best decoders and coverage-map ablation.} In the best configuration (Claude Opus 4.6, detailed prompt), the 92.83\,\% result was obtained by a hybrid method: class-averaged representations serve as centroids, and the decision is based on Mahalanobis distance with class-specific covariance. The pattern is the same for Perplexity Sonar with the detailed prompt: the best decoders belong to LDA/QDA and prototype method classes. These classes use the entire output spike-rate vector, whereas the baseline argmax uses only one of its components. The ablation confirms the role of the coverage map: when it is removed (while the catalog of method classes and Chain-of-Thought are retained), Perplexity Sonar does not reach 90\,\% during a run, and the best accuracy falls from 92.70\,\% to 83.18\,\%; the analyst converges on a narrow set of similar ensemble variations.

The results for Qwen3-30B-A3B-Instruct-2507 are provided in Tables~\ref{tab:qwen_prompt_paired_runs} and~\ref{tab:qwen_ablation_prompt_paired_runs}, those for Perplexity Sonar in Tables~\ref{tab:sonar_prompt_paired_runs} and~\ref{tab:sonar_ablation_prompt_paired_runs}, and those for Claude Opus~4.6 in Tables~\ref{tab:opus_prompt_paired_runs} and~\ref{tab:opus_ablation_prompt_paired_runs}.

\begin{table}[htbp]
\centering
\caption{Maximum accuracy in ten matched pairs of complete agent-system runs with the simple and detailed prompts for Qwen3-30B-A3B-Instruct-2507 on MNIST. Within each pair, the same trained SNN instance and identical output spike data were used. The difference $\Delta$ is calculated as the accuracy for the detailed prompt minus the accuracy for the simple prompt.}
\label{tab:qwen_prompt_paired_runs}
\begin{tabular}{cccc}
\toprule
Pair no. & Simple, \% & Detailed, \% & $\Delta$, pp \\
\midrule
1  & 79.88 & 82.28 & 2.40 \\
2  & 79.88 & 83.35 & 3.47 \\
3  & 79.88 & 88.57 & 8.69 \\
4  & 79.88 & 86.48 & 6.60 \\
5  & 79.88 & 89.16 & 9.28 \\
6  & 79.88 & 84.10 & 4.22 \\
7  & 79.88 & 87.20 & 7.32 \\
8  & 79.88 & 85.50 & 5.62 \\
9  & 79.88 & 83.90 & 4.02 \\
10 & 79.88 & 88.30 & 8.42 \\
\midrule
Mean & 79.88 & 85.88 & 6.00 \\
Std. dev. & 0.00 & 2.42 & 2.42 \\
\bottomrule
\end{tabular}
\end{table}

\begin{table}[htbp]
\centering
\caption{Comparison of maximum accuracy in ten matched Qwen3-30B-A3B-Instruct-2507 runs on MNIST for the detailed prompt with and without the coverage map. Within each pair, the same SNN instance and identical output spike data were used. The difference $\Delta$ is calculated as the accuracy with the coverage map minus the accuracy without the coverage map.}
\label{tab:qwen_ablation_prompt_paired_runs}
\begin{tabular}{cccc}
\toprule
Pair no. & With map, \% & Without map, \% & $\Delta$, pp \\
\midrule
1  & 82.28 & 79.88 & 2.40 \\
2  & 83.35 & 80.42 & 2.93 \\
3  & 88.57 & 79.88 & 8.69 \\
4  & 86.48 & 81.15 & 5.33 \\
5  & 89.16 & 79.88 & 9.28 \\
6  & 84.10 & 80.73 & 3.37 \\
7  & 87.20 & 79.88 & 7.32 \\
8  & 85.50 & 79.88 & 5.62 \\
9  & 83.90 & 80.25 & 3.65 \\
10 & 88.30 & 79.88 & 8.42 \\
\midrule
Mean & 85.88 & 80.18 & 5.70 \\
Std. dev. & 2.42 & 0.45 & 2.58 \\
\bottomrule
\end{tabular}
\end{table}

\begin{table}[htbp]
\centering
\caption{Maximum accuracy in ten matched pairs of complete agent-system runs with the simple and detailed prompts for Perplexity Sonar on MNIST. Within each pair, the same trained SNN instance and identical output spike data were used. The difference $\Delta$ is calculated as the accuracy for the detailed prompt minus the accuracy for the simple prompt.}
\label{tab:sonar_prompt_paired_runs}
\begin{tabular}{cccc}
\toprule
Pair no. & Simple, \% & Detailed, \% & $\Delta$, pp \\
\midrule
1  & 79.88 & 92.70 & 12.82 \\
2  & 79.88 & 92.57 & 12.69 \\
3  & 79.88 & 91.95 & 12.07 \\
4  & 79.88 & 91.82 & 11.94 \\
5  & 79.88 & 91.77 & 11.89 \\
6  & 79.88 & 91.74 & 11.86 \\
7  & 79.88 & 91.64 & 11.76 \\
8  & 79.88 & 91.45 & 11.57 \\
9  & 79.88 & 91.08 & 11.20 \\
10 & 79.88 & 90.93 & 11.05 \\
\midrule
Mean & 79.88 & 91.76 & 11.89 \\
Std. dev. & 0.00 & 0.56 & 0.56 \\
\bottomrule
\end{tabular}
\end{table}

\begin{table}[htbp]
\centering
\caption{Comparison of maximum accuracy in ten matched Perplexity Sonar runs on MNIST for the detailed prompt with and without the coverage map. Within each pair, the same SNN instance and identical output spike data were used. The difference $\Delta$ is calculated as the accuracy with the coverage map minus the accuracy without the coverage map.}
\label{tab:sonar_ablation_prompt_paired_runs}
\begin{tabular}{cccc}
\toprule
Pair no. & With map, \% & Without map, \% & $\Delta$, pp \\
\midrule
1  & 92.70 & 83.18 &  9.52 \\
2  & 92.57 & 82.92 &  9.65 \\
3  & 91.95 & 82.77 &  9.18 \\
4  & 91.82 & 82.66 &  9.16 \\
5  & 91.77 & 82.65 &  9.12 \\
6  & 91.74 & 82.47 &  9.27 \\
7  & 91.64 & 82.15 &  9.49 \\
8  & 91.45 & 82.04 &  9.41 \\
9  & 91.08 & 81.87 &  9.21 \\
10 & 90.93 & 80.71 & 10.22 \\
\midrule
Mean & 91.76 & 82.34 & 9.42 \\
Std. dev. & 0.56 & 0.70 & 0.33 \\
\bottomrule
\end{tabular}
\end{table}

\begin{table}[htbp]
\centering
\caption{Maximum accuracy in ten matched pairs of complete agent-system runs with the simple and detailed prompts for Claude Opus~4.6 on MNIST. Within each pair, the same trained SNN instance and identical output spike data were used. The difference $\Delta$ is calculated as the accuracy for the detailed prompt minus the accuracy for the simple prompt.}
\label{tab:opus_prompt_paired_runs}
\begin{tabular}{cccc}
\toprule
Pair no. & Simple, \% & Detailed, \% & $\Delta$, pp \\
\midrule
1  & 90.20 & 92.83 &  2.63 \\
2  & 91.15 & 91.88 &  0.73 \\
3  & 91.65 & 91.58 & -0.07 \\
4  & 91.60 & 92.49 &  0.89 \\
5  & 91.33 & 91.22 & -0.11 \\
6  & 90.71 & 92.35 &  1.64 \\
7  & 91.42 & 92.69 &  1.27 \\
8  & 91.17 & 92.57 &  1.40 \\
9  & 90.88 & 92.66 &  1.78 \\
10 & 91.22 & 91.24 &  0.02 \\
\midrule
Mean      & 91.13 & 92.15 & 1.02 \\
Std. dev. & 0.44  & 0.62  & 0.90 \\
\bottomrule
\end{tabular}
\end{table}

\begin{table}[htbp]
\centering
\caption{Comparison of maximum accuracy in ten matched Claude Opus~4.6 runs on MNIST for the detailed prompt with and without the coverage map. Within each pair, the same SNN instance and identical output spike data were used. The difference $\Delta$ is calculated as the accuracy with the coverage map minus the accuracy without the coverage map.}
\label{tab:opus_ablation_prompt_paired_runs}
\begin{tabular}{cccc}
\toprule
Pair no. & With map, \% & Without map, \% & $\Delta$, pp \\
\midrule
1  & 91.22 & 90.66 & 0.56 \\
2  & 91.24 & 90.73 & 0.51 \\
3  & 91.58 & 90.89 & 0.69 \\
4  & 91.88 & 90.91 & 0.97 \\
5  & 92.35 & 90.95 & 1.40 \\
6  & 92.49 & 91.01 & 1.48 \\
7  & 92.57 & 91.04 & 1.53 \\
8  & 92.66 & 91.20 & 1.46 \\
9  & 92.69 & 91.32 & 1.37 \\
10 & 92.83 & 91.45 & 1.38 \\
\midrule
Mean & 92.15 & 91.02 & 1.14 \\
Std. dev. & 0.62 & 0.25 & 0.41 \\
\bottomrule
\end{tabular}
\end{table}

When the simple and detailed prompts were compared, the mean maximum accuracy for Qwen3-30B-A3B-Instruct-2507 increased from \(79.88\pm0.00\,\%\) to \(85.88\pm2.42\,\%\). A positive difference was also observed in all ten pairs and averaged \(6.00\) percentage points. For Claude Opus~4.6, the mean maximum increased from \(91.13\pm0.44\,\%\) to \(92.15\pm0.62\,\%\). The mean paired difference was \(1.02\) percentage points: a positive effect was observed in 8 of 10 pairs, whereas accuracy decreased by \(0.07\) and \(0.11\) percentage points in two pairs. For Perplexity Sonar, switching to the detailed prompt increased the mean maximum accuracy from \(79.88\pm0.00\,\%\) to \(91.76\pm0.56\,\%\). A positive difference was observed in all ten pairs, with a mean of \(11.89\) percentage points.

The coverage-map ablation showed an additional effect of this component of the detailed prompt. For Qwen3-30B-A3B-Instruct-2507, mean accuracy increased from \(80.18\pm0.45\,\%\) to \(85.88\pm2.42\,\%\), i.e., by \(5.70\) percentage points. In each of the three ablation comparisons, a positive difference was observed in all ten pairs. For Perplexity Sonar, the corresponding values were \(82.34\pm0.70\,\%\) without the map and \(91.76\pm0.56\,\%\) with the map, with a mean increase of \(9.42\) percentage points. For Claude Opus~4.6, the presence of the map increased the mean maximum accuracy from \(91.02\pm0.25\,\%\) to \(92.15\pm0.62\,\%\); the mean paired increase was \(1.14\) percentage points.

For each model, the statistical significance of differences between the two configurations was assessed using a two-sided exact Wilcoxon signed-rank test for paired samples. The null hypothesis stated that neither compared configuration had a systematic advantage: the paired accuracy differences were centered at zero. The alternative hypothesis assumed a systematic difference between the configurations. The test was applied under the assumption of a symmetric distribution of paired differences. For the two-sided test, the \(W\) statistic was defined as the smaller of the rank sums for positive and negative nonzero differences. The significance level was set to \(\alpha=0.05\).

For the comparison of the simple and detailed prompts for Claude Opus~4.6, \(W=5\) and \(p=0.0195\) were obtained. For Perplexity Sonar and Qwen3-30B-A3B-Instruct-2507, all ten paired differences were positive. Therefore, in both cases the two-sided exact Wilcoxon signed-rank test gave the same result: \(W=0\), \(p=0.001953\). The null hypothesis was rejected for all three models at the \(\alpha=0.05\) significance level.

Two research questions were tested for the three models: the effect of switching from the simple to the detailed prompt and the effect of including the coverage map in the detailed prompt. Thus, six statistical comparisons were performed, and their results were used to formulate the overall conclusion about the effectiveness of prompt structure.

Testing each hypothesis separately at a significance level of \(\alpha=0.05\) does not preserve this level for the full set of comparisons. Specifically, if the six tests were independent and all six null hypotheses were true, the probability of at least one false-positive result would be
\[
1-(1-\alpha)^6
=1-0.95^6
=0.2649,
\]
i.e., approximately \(26.49\,\%\).

To ensure control of the family-wise probability of at least one type I error at \(0.05\), the Holm--Bonferroni correction was applied. This procedure controls the family-wise error even when the statistical tests are dependent. The original \(p\) values were ordered in ascending order and sequentially compared with the thresholds
\[
\frac{\alpha}{6},\;
\frac{\alpha}{5},\;
\ldots,\;
\alpha.
\]

For the five comparisons with an original \(p=0.001953\), the adjusted value was \(p_{\mathrm{Holm}}=0.0117\). For the comparison of the simple and detailed prompts for Claude Opus~4.6, with an original \(p=0.0195\), \(p_{\mathrm{Holm}}=0.0195\) was obtained. All six adjusted \(p\) values were below \(0.05\), so the corresponding null hypotheses were also rejected after correction for multiple comparisons.

When the detailed prompt with and without the coverage map was compared, all ten paired differences were positive for each of the three models. The same values, \(W=0\), \(p=0.001953\). After applying the Holm--Bonferroni correction across all six comparisons, the adjusted value was \(p_{\mathrm{Holm}}=0.0117\). Consequently, removing the coverage map led to a statistically significant decrease in the maximum accuracy achieved for each model.

After correction for multiple comparisons, the results of all six paired tests remained statistically significant. When switching from the simple to the detailed prompt, the mean accuracy increase was \(1.02\) percentage points for Claude Opus~4.6, \(11.89\) percentage points for Perplexity Sonar, and \(6.00\) percentage points for Qwen3-30B-A3B-Instruct-2507. For Claude Opus~4.6, the two-sided exact Wilcoxon signed-rank test gave \(W=5\), an original \(p=0.0195\), and \(p_{\mathrm{Holm}}=0.0195\). For Perplexity Sonar and Qwen3-30B-A3B-Instruct-2507, positive differences were obtained in all ten pairs; for each model, \(W=0\), the original \(p=0.001953\), and \(p_{\mathrm{Holm}}=0.0117\).

\paragraph{Results on the \texttt{digits} dataset.} The same framework was applied to \texttt{digits} \(8\times8\); the run design was the same as for MNIST. The baseline argmax decoder gives an accuracy of 90.83\,\%, whereas the best \(\langle\)Claude Opus 4.6, detailed prompt\(\rangle\) configuration increases it to 96.67\,\% (Table~\ref{tab:digits_results}). The best decoder uses a nonparametric generative classifier based on KDE with a Gaussian kernel in a PCA subspace, and the label is determined using the Bayes rule \(\hat{c} = \arg\max_c [\log p(f \mid c) + \log p(c)]\); hyperparameters were selected using 5-fold cross-validation.

\begin{table}[htbp]
\centering
\caption{Comparison of analyst models and prompt variants on the \texttt{digits} \(8\times8\) dataset. For each configuration, the table reports the number of complete agent runs, the mean and standard deviation of the maximum accuracy achieved within an individual run, the highest accuracy across all runs, and the number of runs in which the 93\,\% threshold was reached. Ten matched runs of 250 iterations were performed for every configuration; the baseline (argmax decoder) gives an accuracy of 90.83\,\%.}
\label{tab:digits_results}
\resizebox{\linewidth}{!}{
\begin{tabular}{@{\extracolsep{\fill}}llcccc@{}}
\hline
Model &
Prompt &
\(n\) &
\makecell{Mean maximum\\accuracy \(\pm\) std, \%} &
\makecell{Best\\accuracy, \%} &
\makecell{Runs with\\accuracy \(\geq93\,\%\)} \\
\hline
Claude Opus~4.6
& simple
& 10
& \(95.07\pm0.52\)
& 95.56
& 10 out of 10 \\
Claude Opus~4.6
& detailed
& 10
& \(96.12\pm0.66\)
& \textbf{96.67}
& 10 out of 10 \\
\hline
Perplexity Sonar
& simple
& 10
& \(93.44\pm0.57\)
& 93.89
& 9 out of 10 \\
Perplexity Sonar
& detailed
& 10
& \(93.31\pm0.93\)
& 94.17
& 8 out of 10 \\
\hline
Qwen3-30B-A3B-Instruct-2507
& simple
& 10
& \(91.63\pm0.74\)
& 92.78
& 0 out of 10 \\
Qwen3-30B-A3B-Instruct-2507
& detailed
& 10
& \(93.66\pm0.79\)
& 94.17
& 7 out of 10 \\
\hline
\end{tabular}
}
\end{table}

\subsection{Decoder search dynamics}
\label{subsec:decoder_search_dynamics}

\begin{figure}
    \centering

    \subcaptionbox{Simple prompt, Qwen3-30B\label{fig:mnist_qwen_simple}}{
        \includegraphics[width=0.32\textwidth]{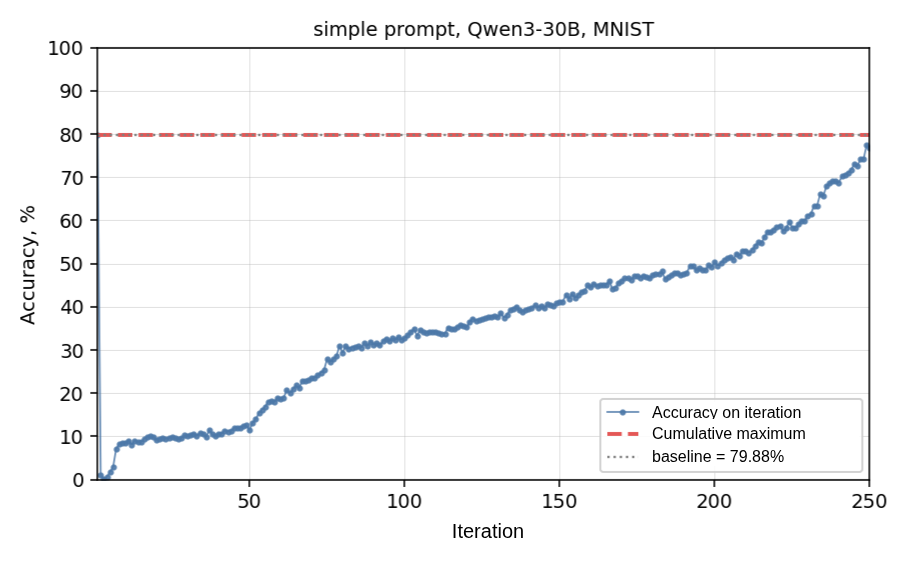}
    }\hfill
    \subcaptionbox{Simple prompt, Perplexity Sonar\label{fig:mnist_sonar_simple}}{
        \includegraphics[width=0.32\textwidth]{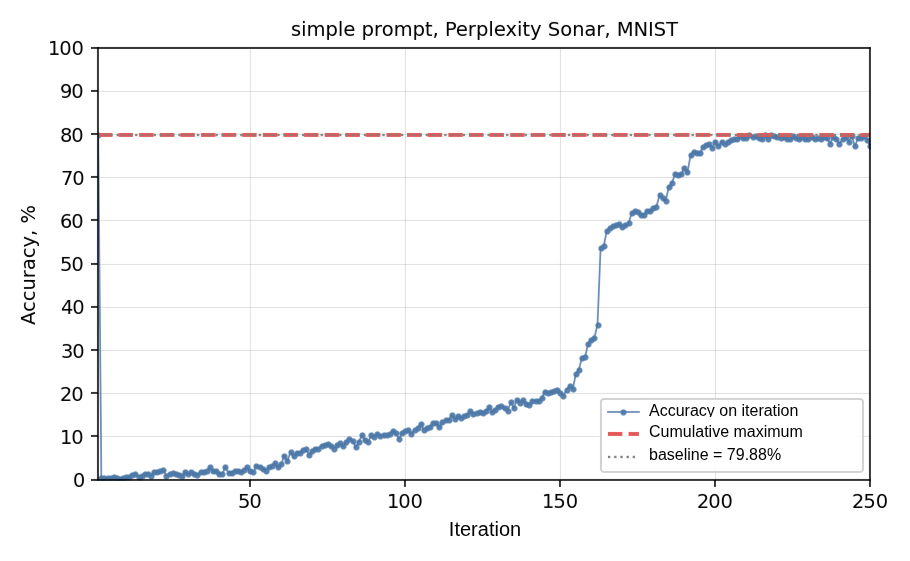}
    }\hfill
    \subcaptionbox{Simple prompt, Claude Opus 4.6\label{fig:mnist_opus_simple}}{
        \includegraphics[width=0.32\textwidth]{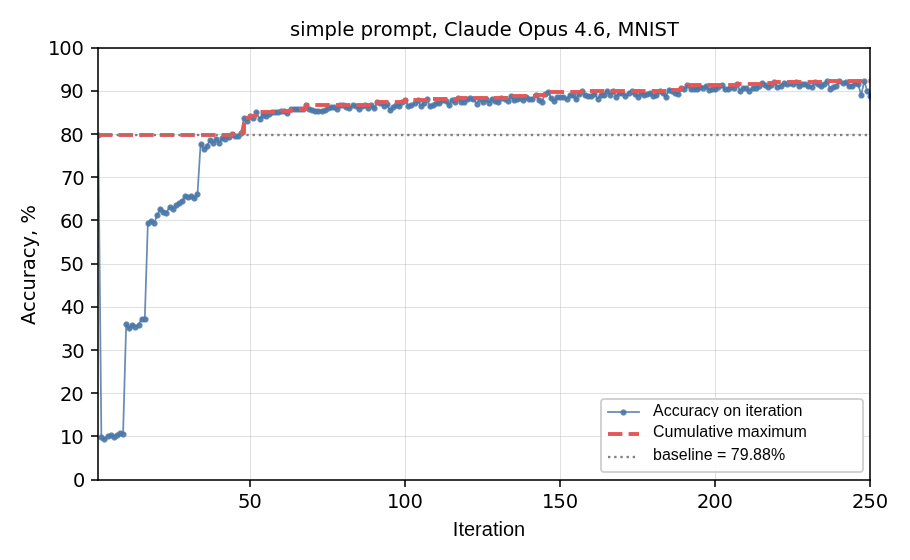}
    }

    \medskip

    \subcaptionbox{Detailed prompt, Qwen3-30B\label{fig:mnist_qwen_detailed}}{
        \includegraphics[width=0.32\textwidth]{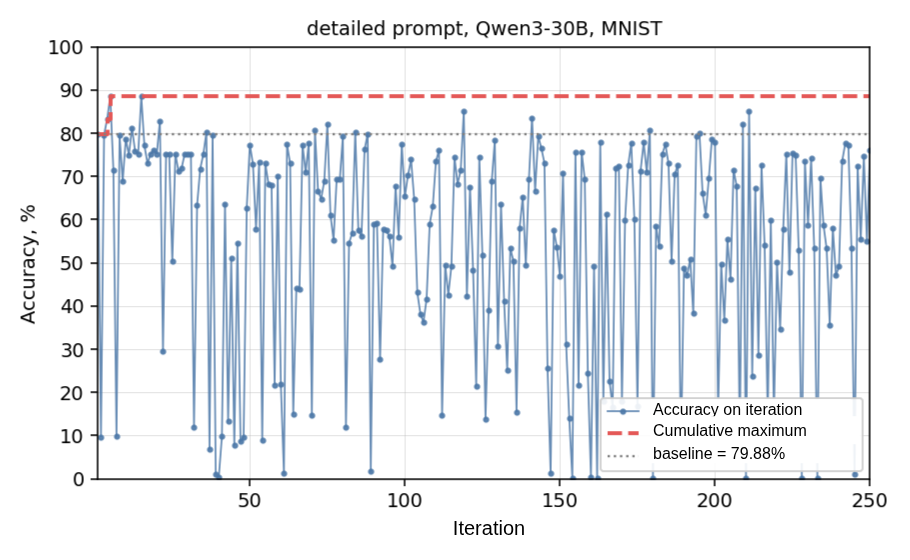}
    }\hfill
    \subcaptionbox{Detailed prompt, Perplexity Sonar\label{fig:mnist_sonar_detailed}}{
        \includegraphics[width=0.32\textwidth]{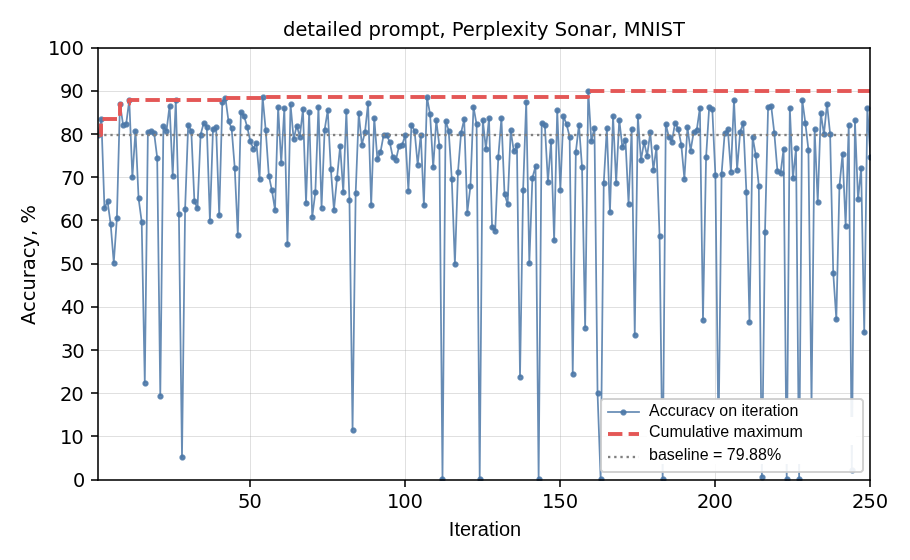}
    }\hfill
    \subcaptionbox{Detailed prompt, Claude Opus 4.6\label{fig:mnist_opus_detailed}}{
        \includegraphics[width=0.32\textwidth]{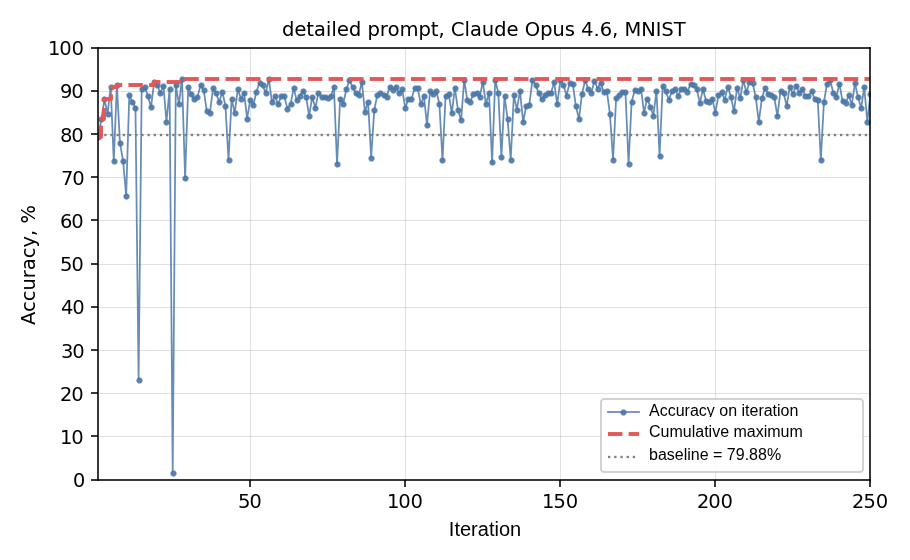}
    }

    \caption{
        Search dynamics for the decoding module on the MNIST dataset.
        The blue line shows the accuracy of the solution obtained at each
        iteration, the red dashed line shows the best result achieved,
        and the gray horizontal line shows the baseline accuracy of 79.88\,\%.
        The columns correspond to analyst models, and the upper and lower rows
        correspond to the simple and detailed prompts, respectively.
    }
    \label{fig:decoder_search_grid}
    \label{fig:decoder_search_mnist}
\end{figure}

\begin{figure}
    \centering

    \subcaptionbox{Simple prompt, Qwen3-30B\label{fig:digits_qwen_simple}}{
        \includegraphics[width=0.32\textwidth]{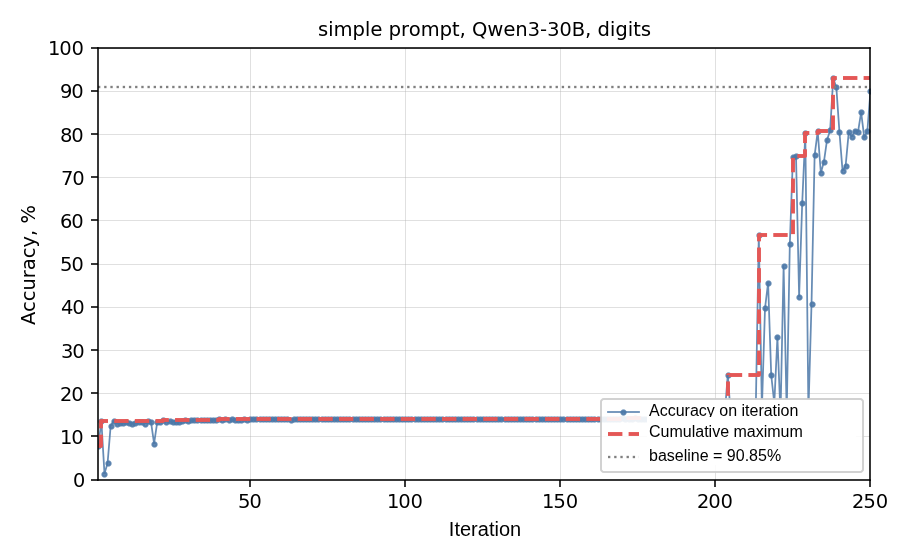}
    }\hfill
    \subcaptionbox{Simple prompt, Perplexity Sonar\label{fig:digits_sonar_simple}}{
        \includegraphics[width=0.32\textwidth]{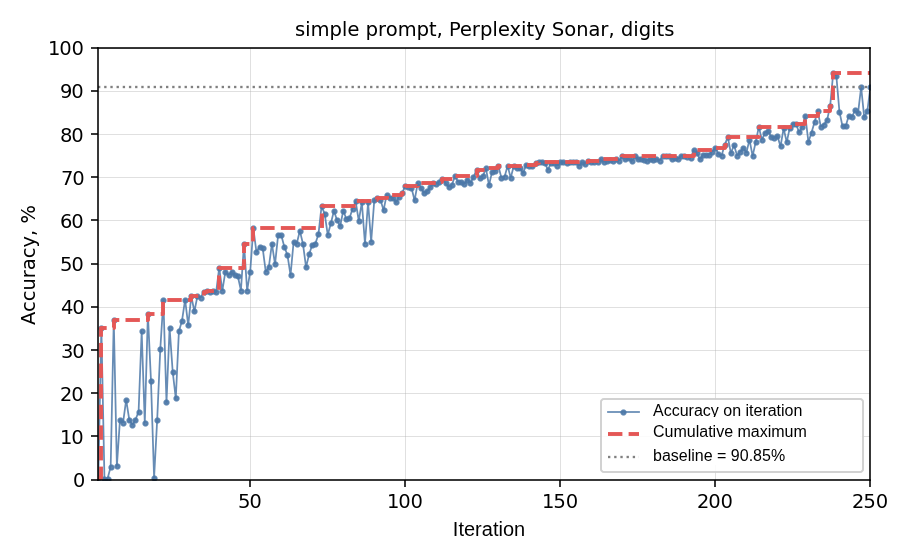}
    }\hfill
    \subcaptionbox{Simple prompt, Claude Opus 4.6\label{fig:digits_opus_simple}}{
        \includegraphics[width=0.32\textwidth]{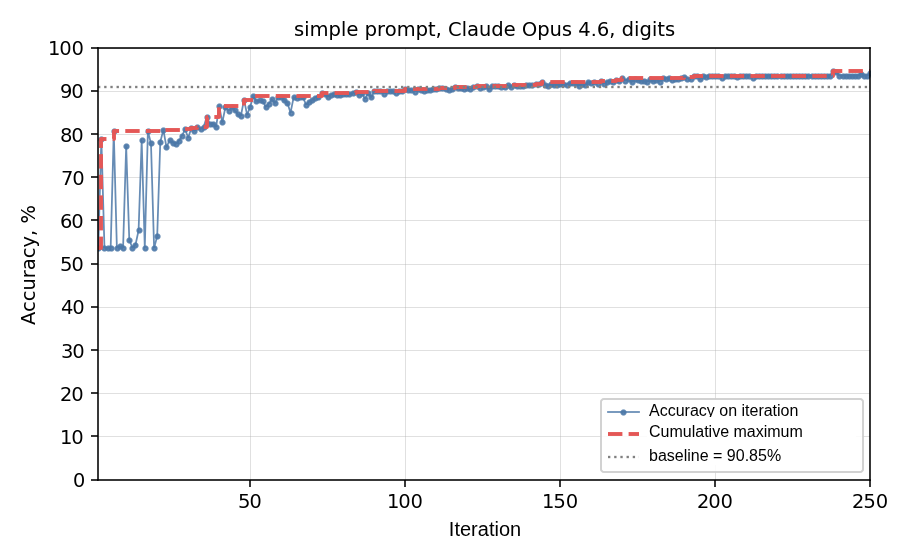}
    }

    \medskip

    \subcaptionbox{Detailed prompt, Qwen3-30B\label{fig:digits_qwen_detailed}}{
        \includegraphics[width=0.32\textwidth]{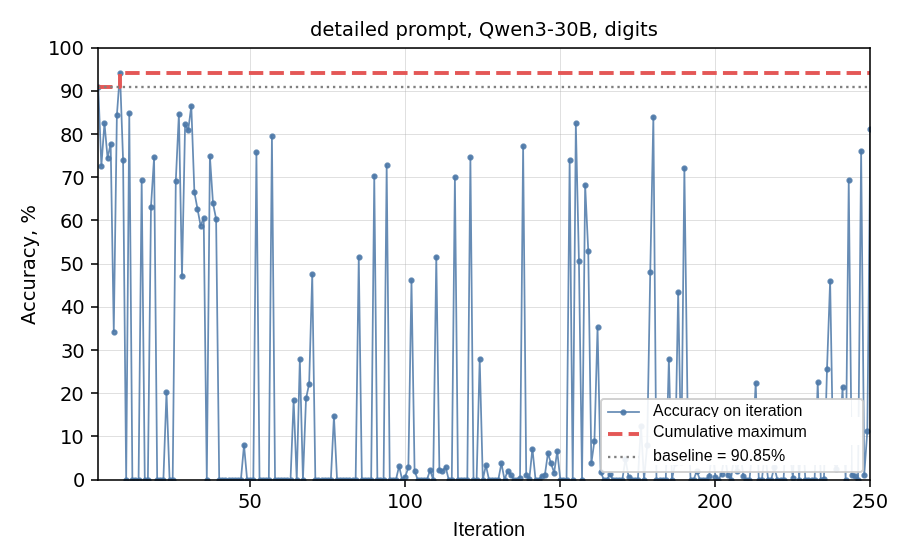}
    }\hfill
    \subcaptionbox{Detailed prompt, Perplexity Sonar\label{fig:digits_sonar_detailed}}{
        \includegraphics[width=0.32\textwidth]{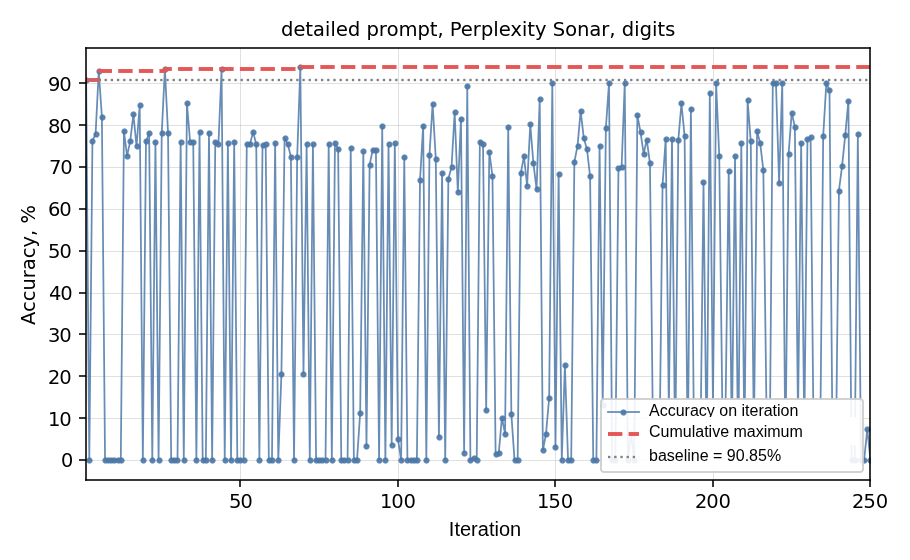}
    }\hfill
    \subcaptionbox{Detailed prompt, Claude Opus 4.6\label{fig:digits_opus_detailed}}{
        \includegraphics[width=0.32\textwidth]{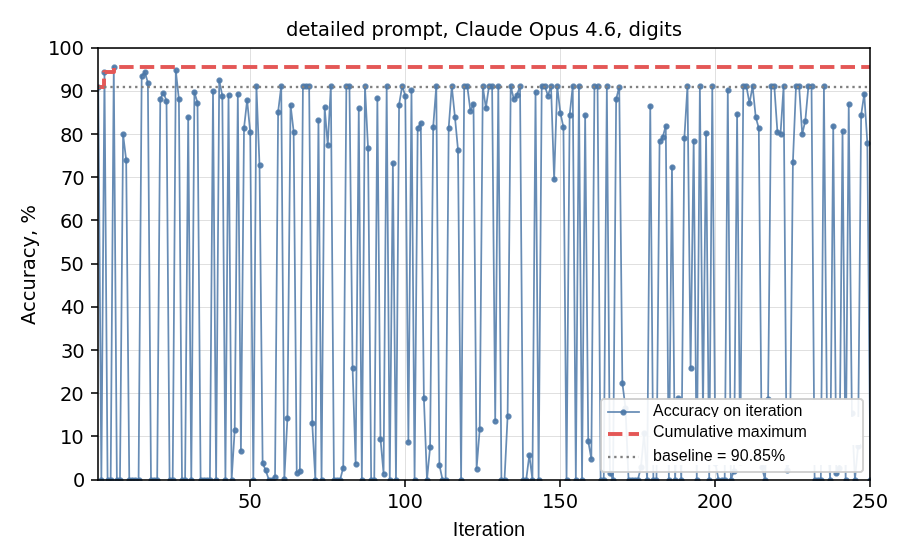}
    }

    \caption{
        Search dynamics for the decoding module on the \texttt{digits} dataset.
        The blue line shows the accuracy of the solution obtained at each
        iteration, the red dashed line shows the best result achieved,
        and the gray horizontal line shows the baseline accuracy of 90.83\,\%.
        The columns correspond to analyst models, and the upper and lower rows
        correspond to the simple and detailed prompts, respectively.
    }
    \label{fig:decoder_search_digits}
\end{figure}

In addition to the final accuracy values reported in Tables~\ref{tab:prompt_model_configs} and~\ref{tab:digits_results}, the search trajectory itself is of interest: how quickly the agent system finds a decoder that outperforms the baseline and reaches a plateau. Figure~\ref{fig:decoder_search_grid} shows the maximum accuracy among all decoders proposed up to the current iteration (the cumulative maximum) for each run of the agent system. The MNIST trajectories are collected in Fig.~\ref{fig:decoder_search_mnist}, and those for the \texttt{digits} dataset in Fig.~\ref{fig:decoder_search_digits}.

The curves demonstrate a pronounced dependence of convergence speed on the analyst model: with Claude Opus~4.6, the system reaches a plateau by iterations 5--10 (Fig.~\ref{fig:decoder_search_grid}, Claude Opus 4.6 panels), whereas Perplexity Sonar in detailed-prompt mode exhibits stepwise dynamics with accuracy gains at iterations 50, 120, and 170; the system periodically returns to revising hypotheses after long plateaus.

For the simple prompt in the Perplexity Sonar and Qwen3-30B-A3B-Instruct-2507 configurations (MNIST), the cumulative maximum does not rise above the baseline level over the entire range of 250 iterations: in these runs, the analyst proposed no decoder that outperformed argmax over aggregated spikes. This contrasts with the detailed-prompt version of the same models, where accuracy begins to increase in the first iterations, indicating the sensitivity of the agent system to the level of detail in the instruction provided to the analyst.

The largest final gain over the baseline was obtained in the detailed-prompt configuration with Claude Opus~4.6 on \texttt{digits} (best-run accuracy $0.9654$ versus the \texttt{digits} baseline $0.9085$; Table~\ref{tab:digits_results}), with the increase occurring almost entirely during the first 10 iterations, while the system spends the remaining more than 200 iterations confirming the maximum found without further improvement.

\subsection{Decoder catalog and TPOT configuration}
\label{app:random_catalog}

The same configuration was used in the \texttt{classic} and \texttt{sequential} TPOT modes, which differ in the search strategy but not in the space of admissible solutions, and in the Random Choice sampler described in Section~\ref{sec:setup_search_baselines_snn}. The \texttt{classic} mode used a population of 25 and 9 generations, the \texttt{sequential} mode a population of 1 and 249 consecutive generations. The budget of each mode was 250 evaluated candidates per run. Random Choice does not use TPOT as a search strategy: it draws configurations from the same catalog through independent calls to Qwen3-30B-A3B-Instruct-2507 and issues as many calls as are needed to record 250 evaluated decoders. Per run it made 347 calls on average.

TPOT selected the best pipeline by mean accuracy in five-fold cross-validation on the training set. After the search was completed, the selected pipeline was retrained on the entire training set and evaluated once on the test set. Test data were not involved in selecting the pipeline structure or its hyperparameters.

\paragraph{Mapping between method classes and operators.}

The catalog is based on the decoder taxonomy from the analyst's detailed prompt. Some substantive classes map to the same scikit-learn estimator with different metrics or hyperparameters. Therefore, the number of TPOT operators does not match the number of classes in the taxonomy. For convenience, the mapping between decoder method classes and TPOT operators is provided in Table~\ref{tab:tpot_catalog_mapping}.

\begin{table}[htbp]
\centering
\caption{Mapping of decoder method classes to TPOT catalog operators.}
\label{tab:tpot_catalog_mapping}
\small
\setlength{\tabcolsep}{5pt}
\renewcommand{\arraystretch}{1.15}
\begin{tabularx}{\linewidth}{
    >{\raggedright\arraybackslash}p{0.26\linewidth}
    >{\raggedright\arraybackslash}p{0.31\linewidth}
    >{\raggedright\arraybackslash}X}
\toprule
Method class &
Catalog operator &
Implementation \\
\midrule

Prototypes and centroids &
\texttt{NearestCentroid} &
Classification by distance to the class centroid \\
\midrule

\(k\)-nearest neighbors &
\texttt{KNeighborsClassifier} &
The number of neighbors, weighting rule, and metric are selected from the grid \\
\midrule

Correlation, cosine, and local-density rules &
\texttt{KNeighborsClassifier} &
Implemented using different local-proximity metrics \\
\midrule

Covariance methods, LDA, and QDA &
\makecell[l]{
    \texttt{LinearDiscriminant}\\[-1pt]
    \texttt{Analysis},\\
    \texttt{QuadraticDiscriminant}\\[-1pt]
    \texttt{Analysis}
} &
Linear and quadratic discriminant rules \\
\midrule

Naive Bayes classifier &
\texttt{GaussianNB} &
Gaussian class-conditional feature model \\
\midrule

Threshold and tree-based rules &
\makecell[l]{
    \texttt{DecisionTreeClassifier},\\
    \texttt{ExtraTreeClassifier}
} &
Threshold partitions of the spike-rate space \\
\midrule

Tree ensembles &
\makecell[l]{
    \texttt{RandomForestClassifier},\\
    \texttt{ExtraTreesClassifier}
} &
Voting by an ensemble of threshold decoders \\
\midrule

Linear class scores &
\texttt{LogisticRegression} &
Linear decision rule without changing the SNN parameters \\
\midrule

Method combinations &
\makecell[l]{
    \texttt{StackingEstimator},\\
    \texttt{VotingClassifier}
} &
Multistage pipelines and prediction aggregation \\
\bottomrule
\end{tabularx}
\end{table}

\paragraph{Hyperparameter grids.}
The complete search-space configuration is provided below. Parameter
values were selected only from the listed discrete sets.

\begin{Verbatim}[
    breaklines=true,
    breakanywhere=true,
    fontsize=\scriptsize
]
CONFIG_DETAILED_WITH_ENSEMBLE = {
    "sklearn.neighbors.NearestCentroid": {
        "metric": ["euclidean", "manhattan"],
    },

    "sklearn.neighbors.KNeighborsClassifier": {
        "n_neighbors": range(1, 31),
        "weights": ["uniform", "distance"],
        "metric": [
            "euclidean",
            "manhattan",
            "cosine",
            "chebyshev",
        ],
    },

    "sklearn.discriminant_analysis."
    "LinearDiscriminantAnalysis": {
        "solver": ["svd", "lsqr", "eigen"],
        "shrinkage": [None, "auto"],
    },

    "sklearn.discriminant_analysis."
    "QuadraticDiscriminantAnalysis": {
        "reg_param": [0.0, 1e-3, 1e-2, 1e-1, 0.5],
    },

    "sklearn.naive_bayes.GaussianNB": {},

    "sklearn.tree.DecisionTreeClassifier": {
        "criterion": ["gini", "entropy"],
        "max_depth": range(1, 11),
        "min_samples_split": range(2, 21),
        "min_samples_leaf": range(1, 21),
    },

    "sklearn.tree.ExtraTreeClassifier": {
        "criterion": ["gini", "entropy"],
        "max_depth": range(1, 11),
        "min_samples_split": range(2, 21),
        "min_samples_leaf": range(1, 21),
    },

    "sklearn.ensemble.RandomForestClassifier": {
        "n_estimators": [100, 200],
        "criterion": ["gini", "entropy"],
        "max_features": [0.5, 0.75, 1.0, "sqrt"],
        "min_samples_split": range(2, 11),
        "min_samples_leaf": range(1, 11),
        "bootstrap": [True, False],
    },

    "sklearn.ensemble.ExtraTreesClassifier": {
        "n_estimators": [100, 200],
        "criterion": ["gini", "entropy"],
        "max_features": [0.5, 0.75, 1.0, "sqrt"],
        "min_samples_split": range(2, 11),
        "min_samples_leaf": range(1, 11),
        "bootstrap": [True, False],
    },

    "sklearn.linear_model.LogisticRegression": {
        "penalty": ["l2"],
        "C": [0.01, 0.1, 1.0, 10.0, 100.0],
        "solver": ["lbfgs", "newton-cg"],
        "max_iter": [1000],
    },

    "tpot.builtins.StackingEstimator": {
        "estimator": [
            "sklearn.neighbors.NearestCentroid",
            "sklearn.neighbors.KNeighborsClassifier",
            "sklearn.discriminant_analysis."
            "LinearDiscriminantAnalysis",
            "sklearn.discriminant_analysis."
            "QuadraticDiscriminantAnalysis",
            "sklearn.naive_bayes.GaussianNB",
            "sklearn.tree.DecisionTreeClassifier",
            "sklearn.tree.ExtraTreeClassifier",
            "sklearn.ensemble.RandomForestClassifier",
            "sklearn.ensemble.ExtraTreesClassifier",
            "sklearn.linear_model.LogisticRegression",
        ],
    },

    "sklearn.ensemble.VotingClassifier": {
        "voting": ["soft"],
    },
}
\end{Verbatim}

The \texttt{StackingEstimator} and \texttt{VotingClassifier} operators were included in the catalog in all modes.

\section{Multi-Agent Framework Prompts}
\label{app:prompts}

The original language of all prompts is Russian.

\subsection{Analyst prompt (simple)}
\label{suppl:prompt_simple}
\begin{Verbatim}[breaklines=true, breakanywhere=true, fontsize=\footnotesize]
# Context 
You are a leading expert in spiking neural networks (SNNs) and in decoding outputs firing rates into class labels for the MNIST task, with over 20 years of practical experience.
You are provided with: 
- Output spike sequences from 10 output neurons for the training and test datasets; 
- A JSON table containing experiments conducted using various decoding methods. 
The goal of this research is to develop a new decoding method that achieves an accuracy of 0.9 or higher (90%+) using a vector of firing rates from 10 output neurons. 

Important feature of the encoding: 
Input sequences are encoded by randomly selecting spikes without replacement from a given reference spike sequence. Therefore, the temporal structure of individual spike sequences is unstable across runs and should not be used directly. The informative object for decoding is the vector of firing rates of the 10 output neurons. 

# Your task 
Develop a method for decoding the firing-rate vector of the output neurons (dimension: 10) that is not included in the provided JSON experiment table, with the goal of achieving an accuracy of 0.9 or higher. 

**Available data and abilities:** 
1. Output spike sequences vector (zeros and ones) for the training dataset - train_spikes.
2. Output spike sequences vector (zeros and ones) for the test dataset - test_spikes.
3. Correct labels for the training dataset - y_train.
4. You may construct new features and new decoding rules based on firing_rates.

When developing a method, clearly distinguish between:
- the feature-extraction step from the firing rate vector;
- the decision rule step based on these features.

Examples of acceptable method classes (which may be combined or extended, but are not limited to):
- Normalized argmax over firing rates, optionally with neuron-specific weights and biases;
- Rules based on the difference between the two highest firing rates (margin-based);
- Class calibration using firing rate statistics for each class (e.g., mean firing rates for each neuron and class) and calss-specific score calculations;
- Simple threshold-based and piecewise rules (e.g., small if/then/else/ decision trees based on firing rate thresholds and their combinations).

**Limitations:**
1. The output spike trains have already been recorded. You cannot modify the network architecture or encoding method to record different sequences.
2. You cannot use backpropagation to train the proposed method.
3. You may fit method parameters only on the training data (train_spikes and y_train). The test data must be used exclusively for evaluation.
4. Your answer must be presented as a developer specification, including:
 - a clear description of how features are constructed from firing rates (optional);
 - a description of the decision rule in analytical form:
 - formulas for intermediate calculations and for the final transformation from firing rates to a class label;
 - recommended hyperparameters in the following JSON: `{"decoding_params": {}}`,
 where `decoding_params` is a dictionary of parameters, such as thresholds, weights, and search ranges.
5. Do not repeat methods that have already been tested and included in the JSON experiment table.
6. If a parameter is best tuned using the training data (e.g., thresholds, weights, or ranges), mark it with the value `"to_fit"` in `decoding_params` and provide a grid of candidate values for iteration.

# Experiment Table (JSON):
\end{Verbatim}

\subsection{Analyst prompt (detailed)}
\label{suppl:prompt_detailed}

\begin{Verbatim}[breaklines=true, breakanywhere=true, fontsize=\footnotesize]
You are a leading neuroinformatics expert (20+ years of experience in spiking neural networks (SNNs) and in decoding outputs firing rates into class labels.
Your specialty is developing decoding methods that operate WITHOUT backpropagation, relying exclusively on the statistical properties of output-neuron firing rates.

# Dataset: {dataset_name}
- Task: {task_description}
- Output neurons: (n_neurons)
- Classes: {n_classes}
- Class labels: {class_labels}
- **Target accuracy: {target_accuracy}+**

## Input data
- `train_spikes` - a NumPy array `(N_train, {n_neurons})`: the firing rates of the output neurons for the training set;
- `test_spikes` - a NumPy array `(N_test, {n_neurons}0`: the firing rates of the output neurons for the test set;
- `y_train` - a NumPy array `(N_train,)`: true class labels.

## Encoding Feature
Input sequences are generated by randomly sampling spikes from a refrence sequence.
Their temporal structure is **unstable** across runs.
**Only** the firing rate vector of the {n_neurons} output neurons is informative.

# Current experiments on the {dataset_name} Dataset
A total of {n_experiments} experiments were conducted. Below is the table (JSON):

```json
{current_results_json}
```

# Method Class Coverage
Below is an automated analysis of which **method classes** have already been tested and which remain unexplored. 
Your task is to **prioritize methods from unexplored or underexplored classes** whenever they are theoretically applicable to the given problem.

{method_class_coverage_section}

**Method class selection rule:**
If the table above contains method classes with the status "NOT STUDIED" or "LIMITED DATA", propose a method from one of those classes.
Consider an already well-studied class only if you can justify why the proposed variation is fundamntally different from all previously tested methods.

# Task
Develop a **new** method for decoding firing rate vectors of dimension {n_neurons} that is **not included** in the experiment table, with the goal of achieving an accuracy of **{target_accuracy}+**.

## Mandatory Analytical Process
Complete the Chain-of-Thought reasoning:

### Step 1 - Class Coverage Analysis
Review the method class coverage table above.
- Which classes have not been studied or remain underexplored?
- Do any of those classes include methods that are theoretically suitable for this task?
- What is the highest accuracy achieved by the methods already tested?
- Which properties such as normalization, distance metrics, or regulatization may have contributed to thier performance?

### Step 2 - Failure Analysis
Which methods degraded to near-chance accuracy (`~{random_accuracy:.2f}`)?
Formulate hypotheses about the possible causes:
- overfitting;
- matrix degeneracy;
- scale instability;
- bugs involving array dimensions.

**Important:** the failure of a specific implementation does not imply that the entire method class has failed. If kNN or Mahalanobis methods were implemented incorrectly - for example, using a singular matrix or improper normalization - this is a reason to propose **corrected** version rather than abandon the class.

### Step 3 - Selecting a class and synthesizing a method
1. **Priority:** Choose a method from the class with the smallest coverage.
2. If you chose a method from a well-studied class, explain how it fundamentally differs from all previous methofs in that class.
3. Use best practices from successful experiments, such as normalization and regularization, but apply them to a **new type of decision rule**.

When developing the method, distinguish between:
- **Feature extraction:** constructing features from firing rates.
- **Decision rule:** assinging a class label based on these features.

### Step 4 - Self-Check (Required)
Before submitting your final answer, verify the following:
- [ ] Is the proposed method different from all methods already tested?
- [ ] Does the method belong to a **different class** from the one used in most previous experiment? If not, explain why this particular class was selected.
- [ ] When working with covariance matrices: was regulatization added (ridge: `Cov + eps * I`, or `np.linalg.pinv`, or diagonal approximation)?
- [ ] Does the method scale to {n_classes} classes?
- [ ] Is the output an array of integers from 0 to {n_classes_minus_1}?

## Method class Catalog
Below is the complete catalog of permitted method classes. Methods from different classes may be combined. **All** methods are implementable using pure NumPy.

### 1. k-Nearest Neighbors (kNN)
For each test example, find the k nearest neighbors in train_spikes and assign the class by voting. Metrics: Euclidean, Manhattan, Chebyshev, and inverse-distance-weighted voted (1/d).
**Tip:** Apply Z-normalization before classification and select k using leave-one-out validation on the training data.

### 2. Mahalanobis Distance
Calculate class centroids and covariance matrices, then assign each example to the class with the minimum Mahalanobis distance. Implementable in Numpy via `np.linalg.inv` (or `np.linalg.pinv` for robustness).
**Tip:** Apply ridge regularization `Cov_c + eps * I` with eps ~ 1e-4...1e-1. You may also use a pooled covariance matrix instead of the class-specific one.

### 3. Naive Bayes (Gaussian NB)
Estimate \mu_c and \sigma_c for each neuron and class. Calculate the log-likelihood of each class as the sum of log-Gaussian probabilites, then select the class with the highest log-posterior using argmax.
**Tip:** Apply variance smoothing: \sigma^2 = max(\sigma^2, \epsilon).

### 4. Discriminant Analysis (LDA / QDA)
LDA: general covariance matrix + linear discriminant functions.
QDA: class-specific covariances. 
Fisher LDA: projection onto the axis of maximum interclass variance. 
All variants can be implemented using NumPy linear algebra.
**Tip:** For LDA, solve the generalized eigenvalue problem S_w^-1 @ S_b for example using `np.linalg.eigh`.

### 5. Prototypes / Centroids
Calculate a centroid for each class, then assign each example to the class with the minimum Euclidean or cosine distance.
Options: L1/L2 normalization, weighted centroids, and
median centroids for robustness.

### 6. Correlation Decoding
Calculate a class-specific score: Score(x, c) = Pearson(x, template_c) or Score(x, c) = cosine(x, template_c).
The class template is the mean or median vector. Assign the class label using Argmax over the class scores.

### 7. KDE / Density Estimation
For each class, estimate the density p(x|c) from the training data using the kernel method (Gaussian kernel). Classify each example based on argmax p(x|c) * p(c).
**Tip:** Select bandwidth using Silverman's rule or by maximazing accuracy on the training data.

### 8. Rank-based methods
Convert firing rates into ranks, work with the order of neurons.
Options: rank-based Z-score, Spearman correlation, margin between ranks.

### 9. Margin-based rules
Differences between top-K firing rates. Score = fr[top1] - mean(fr[rest]).
Options: normalization, Z-score, temperature scaling.

### 10. Softmax Calibration
Softmax with temperature fitted to the training data. Plus: biases by class.
Z-normalization can be added before softmax.

### 11. Threshold/Piecewise-Specific Rules
Cascade threshold rules (mini-trees), if neuron_k > threshold -> class c. Thresholds are selected on the training data. 
Options: binary codes, one-versus-rest thresholds.

### 12. Ensembles of heterogeneous methods
Combine 2-3 **different** methods from the list above with voting or weighted averaging of scores. Select weights on the training data.

# Response format - Technical Assignment (TA)
Please formulate your answer **strictly** in the following structure:

## 1. Method name
A short, unique name (e.g. "ARMD-v2").

## 2. Method class
Indicate which class from the catalog the method belongs to (for example: "kNN,", "Mahalanobis," "Naive Bayes").

## 3. Method description
One paragraph: what the method does, what mathematical idea it is based on.

## 4. Feature extraction
Formulas for constructing features from firing rates:
- What intermediate values are calculated?
- What normalizations/transformations are applied?

## 5. Decision rule
Analytical decision-making formula:
- How is the score for each class derived from features?
- How is the final label derived from scores?

## 6. Hyperparameters
JSON:
```json
{{"decoding_params":{{...}}}}
```

- Fixed parameters -- with specific values.
- Parameters for training -- with the "to_fit" value and a grid specification.

# Strict constraints
1. The spike sequences have already been recorded; you cannot modify the SNN.
2. **Prohibited:** backpropagation, gradient descent.
3. Parameter fiting is permitted **only** on train_spikes and y_train. Test data may be used exclusively for evaluation.
4. Do not repeat the methods from the experiment table.
5. Output: one-dimensional numpy array of integers ({class_labels}), of length N_test.
\end{Verbatim}

\subsection{Coder prompt}
\label{suppl:prompt_coder}
\begin{Verbatim}[breaklines=true, breakanywhere=true, fontsize=\footnotesize]
# Task
The technical specifications are provided below.
Please write Python code for them.

Write a class that decodes output spike sequences into class labels for the {dataset_name} clasification problem.
Decoding must be based on firing rate vectors - i.e., spike counts aggregated over time - from {n_neurons} output neurons.

**The class constructor takes exactly 2 parameters:**
1. train_spikes -- a two-dimensional NumPy array of the output firing rates of neurons in the training set of shape (N_train, 10).
2. y_train -- a one-dimensional NumPy array of the class labels of the training set of length (N_train,).

**What does the __call__() method accept:**
- test_spikes -- a two-dimensional NumPy array of the output firing rates of neurons on the test sample of shape (N_test, 10).

**Output of the __call__() method:**
y_pred is a one-dimensional Numpy array of length N_test

**Restrictions:**
- Only standard Python and NumPy are allowed; no other third-party libraries are allowed.
- The class name is strictly: Decoding.
- The class constructor takes exactly 2 parameters, described above.
- If necessary, you can implement a fit method that doesn't take any arguments and uses only self.train_spikes and self.y_train.
- If there are parameters in decoding_params with "to_fit" values, you must implement their tuning on the training data (for example, by selecting thresholds or weights using a grid or another simple method).
- After completing the tuning, all hyperparameters (including those marked as "to_fit") should be written to the self.decoding_params attribute as a dictionary with specific values.
- The call occurs through the __call__(self, test_spikes) method.
- Slice indices must be strictly integer; float/fractional indices are not allowed in slices.
- The output is a strictly one-dimensional numpy array of integers from 0 to 9 inclusive,
length test_spikes.shape[0].
- By the time the fit method (if implemented) completes, the class must have a self.decoding_params attribute,
containing the final values of all the decoding method's hyperparameters.

**Strict withdrawal requirements:**
- Return only the decoding code without any comments, text, or additional lines.
Blank for this class:
```python
class Decoding:
    def __init__(self, train_spikes, y_train):
        self.train_spikes = train_spikes
        self.y_train = y_train
    def __call__(self, test_spikes):
        pass
        # return y_pred
```
\end{Verbatim}

\subsection{Recorder prompt}
\label{suppl:prompt_protocol}
\begin{Verbatim}[breaklines=true, breakanywhere=true, fontsize=\footnotesize]
Write a short description (approximately one paragraph) of the decoding method and provide the hyperparameters for the decoding method.
Compose a short name for the experiment. Provide your answer as JSON with the following structure:
```json
{
"name": "Short name",
"description": "Description of what the Decoding layer does",
"decoding_params": {}
}
```
\end{Verbatim}

\bibliographystyle{cas-model2-names}
\bibliography{references}

\end{document}